\PassOptionsToPackage{table}{xcolor}
\documentclass[11pt]{article}

\usepackage[final]{acl}

\usepackage{times}
\usepackage{latexsym}

\usepackage[T1]{fontenc}

\usepackage[utf8]{inputenc}

\usepackage{microtype}

\usepackage{inconsolata}

\usepackage{hyperref}       
\usepackage{url}            
\usepackage{booktabs}       
\usepackage{amsfonts}       
\usepackage{nicefrac}       
\usepackage{xspace}
\usepackage{amsmath}
\usepackage{amssymb}
\usepackage{mathtools}
\usepackage{amsthm}
\usepackage{float}          
\usepackage{adjustbox}      
\usepackage{graphicx}       
\usepackage{caption}        
\usepackage{array}          
\usepackage{graphicx}
\usepackage{array}
\usepackage{wrapfig}  
\usepackage{colortbl}

\usepackage{tcolorbox}
\usepackage{multirow} 
\usepackage{graphicx}     
\usepackage{xspace}       
\usepackage{url}          
\usepackage{hyperref}     
\usepackage{enumitem}
\usepackage{listings}     

\usepackage{cleveref}
\crefname{section}{\S}{\S\S}     
\Crefname{section}{\S}{\S\S}     
\usepackage{subcaption}
\usepackage{marvosym} 
\usepackage{pifont}     
\usepackage{makecell}   
\newcommand{\model}{\textsc{Mentor}\xspace}

\title{MENTOR: Efficient Autoregressive Image Generation with Balanced Multimodal Control}

\author{%
    Haozhe Zhao$^{1}$$^\ast$,~~Zefan Cai$^{2}$$^\ast$,~~Shuzheng Si$^{3}$,~~Liang Chen$^{4}$,\\
    ~~\textbf{Jiuxiang Gu},~~\textbf{Wen Xiao}$^{5}$,~~\textbf{Minjia Zhang}$^{1}$,~~\textbf{Junjie Hu}$^{2}$
    \\
    $^{1}$University of Illinois Urbana-Champaign 
$^{2}$University of
Wisconsin-Madison \\
$^{3}$Tsinghua University
$^{4}$Peking University 
$^{5}$Microsoft \\
    \texttt{haozhez6@illinois.edu, \{zefancai, jhu\}@cs.wisc.edu}\\
    \href{https://haozhezhao.github.io/MENTOR.page}{\texttt{haozhezhao.github.io/MENTOR.page}}
}

\begin{document}
\maketitle

\maketitle
\renewcommand{\thefootnote}{\fnsymbol{footnote}}

\begin{abstract}

Recent text-to-image models achieve impressive visual quality but still face challenges in precise controllability, balancing multimodal inputs, and high training cost for multimodal image generation.
To address these limitations, we propose \textbf{\model}, an autoregressive (AR) framework with a two-stage training paradigm for controllable multimodal image generation:
(1) a \textit{multimodal alignment stage} that establishes robust pixel and semantic-level alignment between inputs and generated tokens, followed by (2) a \textit{multimodal instruction tuning stage} that balances the model's integration of multimodal inputs and enhances generation controllability.
Extensive experiments on DreamBench++ and DreamBench demonstrate that, despite modest model size and training resources, \model achieves a strong balance between textual and visual guidance for controllable image generation, delivering competitive performance at significantly lower computational cost compared to leading baselines. Moreover, our approach attains superior image reconstruction fidelity, broad adaptability across different tasks, and training efficiency.
%

\begin{figure}[htbp]
    \centering
    \includegraphics[width=\linewidth]{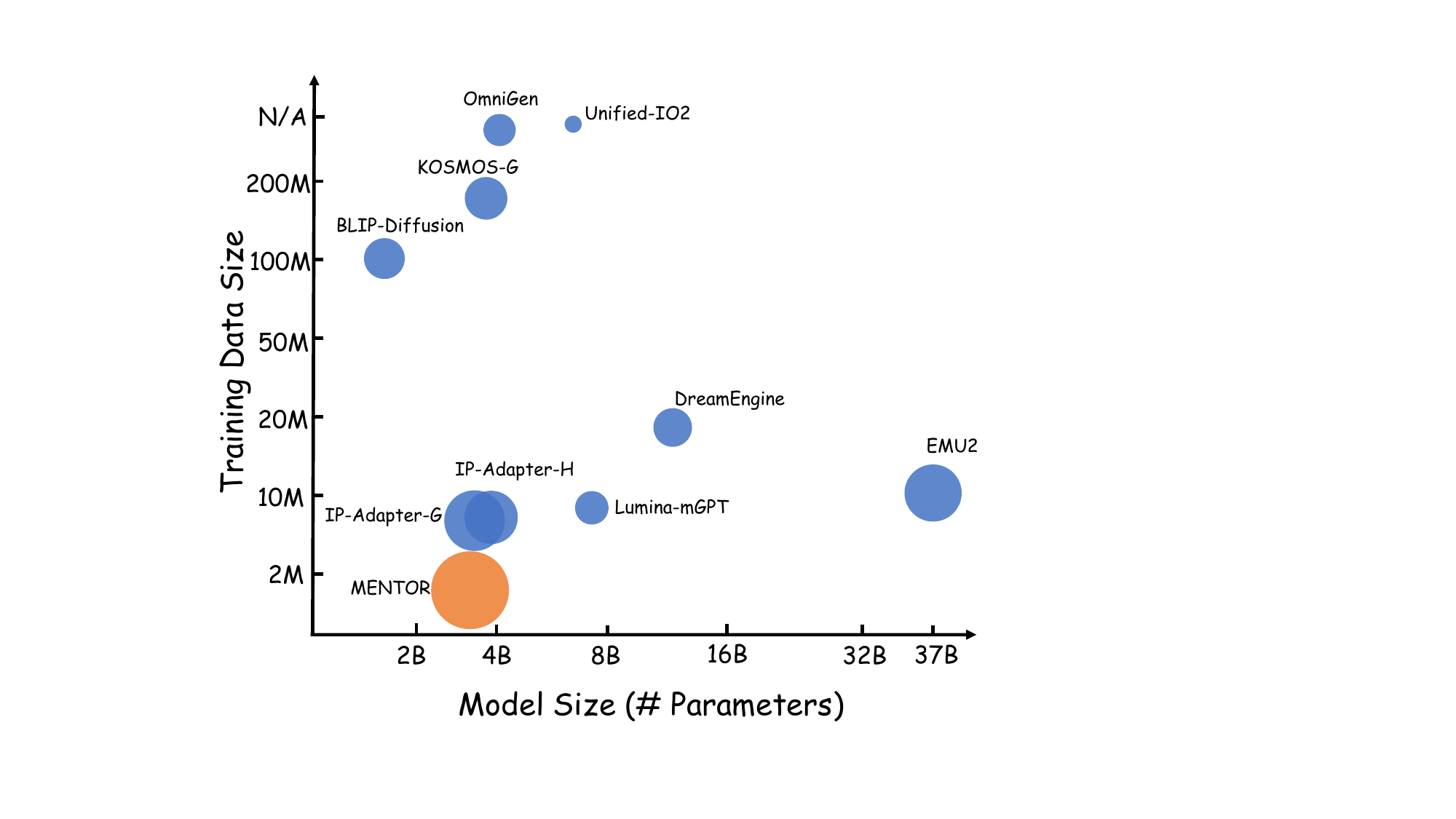}
    \caption{CP$\cdotp$PF score (circle size) of \textcolor[HTML]{ED8B50}{\textbf{\model}} and \textcolor[HTML]{5A76A9}{\textbf{other baselines}} on DreamBench++. Model in lower left achieves the best efficiency.}
    \label{fig:SnapKV}
\end{figure}

\begin{figure*}[htbp]
\centering
\includegraphics[width=0.9\textwidth]{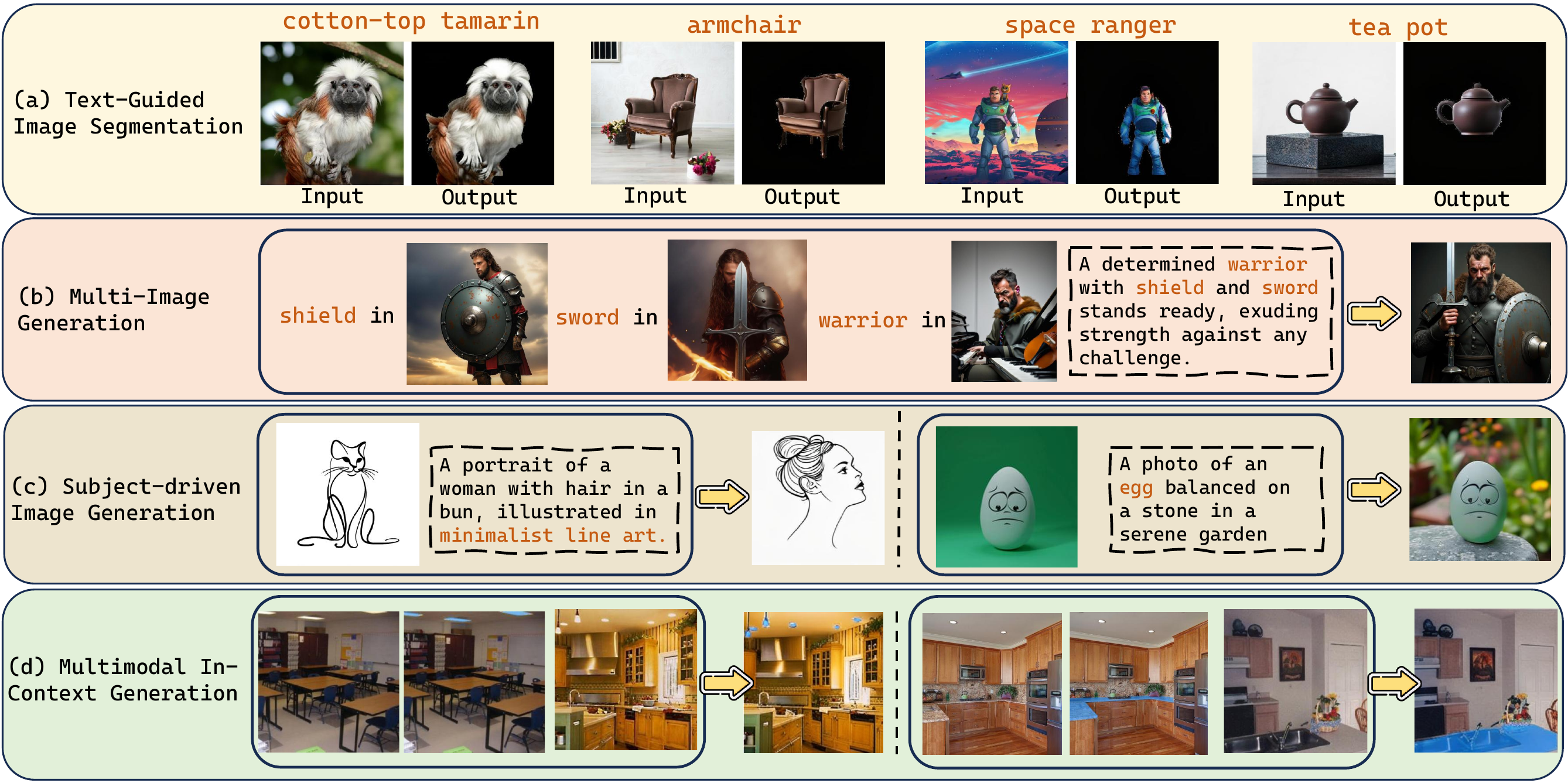}
\caption{Qualitative examples of different tasks built on \model after simply fine-tuning.
}
\label{fig:examples}
\end{figure*}
\end{abstract}

\section{Introduction} 
\label{sec:introduction}

Recent progress in generative models has revolutionized text-to-image (T2I) generation~\citep{2020DDPM,ldm,SDXL}. However, real-world applications often require more than text-only prompts.
To achieve \emph{controllable} image generation, e.g., fine-grained control over generated images, models need to seamlessly integrate multi-modal inputs, such as a reference image together with a detailed text prompt.
This poses significant challenges for existing models that are predominantly focused on T2I generation.
To address this, researchers integrate Large Multimodal Models (LMMs) with generative models \citep{Kosmos-G, emu2,xiao2024omnigenunifiedimagegeneration,zhuang2025vargptv11improvevisualautoregressive,longcatnext} to better handle multimodal inputs. Although effective for tasks like interleaved image-text generation \citep{emu2} and multimodal in-context learning \citep{Kosmos-G, 2024SeedX}, these approaches still face three challenges when scaling to \emph{complex multimodal control}, especially under limited resources:

\textbf{First,} stochastic sampling in diffusion processes makes \emph{precise and deterministic} control difficult, which is essential for high-fidelity tasks (e.g., faithful reconstruction)~\citep{wang2025imageeditingdiffusionmodels}.
\textbf{Second,} balancing guidance across modalities remains challenging. Existing methods exhibit modality imbalance, overemphasize one modality while neglecting the other~\citep{han2024emmatexttoimagediffusionmodel}. This phenomenon appears in both diffusion and autoregressive (AR) paradigms—for instance, IP-Adapter~\citep{ye2023ip-adapter} and Lumina-mGPT~\citep{2024lumina}, conditioned on text and image features, tend to favor image guidance. Such imbalance may stem from modality gaps, architectural limitations \citep{zhao2023mmicl,Dig2DIG,ye2023ip-adapter}, or suboptimal training schemes \citep{Kosmos-G,han2024emmatexttoimagediffusionmodel}.
\textbf{Third,} existing methods rely on auxiliary alignment components, such as learned adapters \citep{Kosmos-G}, regression heads \citep{sun2023emu1,emu2} or specialized embeddings \citep{seed-tokenizer}), and demand large-scale training \citep{emu2,Kosmos-G,2024SeedX}, leading to significant computational costs.
These observations motivate a question: \textit{Is it possible to design an efficient framework for controllable multimodal image generation under limited resources?}

To address these limitations, we propose \textbf{\model}, an efficient autoregressive (AR) framework for controllable multimodal image generation. Unlike diffusion models that demand complex cross-attention layers for multimodal conditioning and extensive training resources \citep{emu2,Kosmos-G,li2023blipdiffusionpretrainedsubjectrepresentation},  \model\ employs a unified transformer that directly aligns multimodal inputs with output tokens, simplifying the architecture, removing auxiliary alignment modules, and substantially reducing training costs.
Our framework uses a multimodal encoder to project inputs into a unified representation, which a transformer decoder then uses to deterministically generate image tokens. To ensure effective and balanced modality integration \citep{han2024emmatexttoimagediffusionmodel}, we further adopt a two-stage training paradigm: (1) a \textit{multimodal alignment stage} that builds robust pixel- and semantic-level alignment between inputs and generated tokens, followed by (2) a \textit{multimodal instruction tuning stage} that balances the modality fusion and enhances generation controllability.

Notably, despite its simplicity and usage of suboptimal checkpoints, \model achieves competitive performance on various benchmarks including Dreambench~\citep{ruiz2023dreamboothfinetuningtexttoimage} and Dreambench++~\citep{peng2025dreambenchpp} with over 10× less resources than leading baselines.
It surpasses resource-intensive models equipped with powerful generators such as SDXL~\citep{SDXL}, SD3~\citep{2024SD3} and Infinity~\citep{han2024infinityscalingbitwiseautoregressive}. 
Controlled experiments show that \model offers a favorable trade-off between efficiency and multimodal balance.
Further analyses highlight its adaptability across reinforcement learning and diverse multimodal tasks, offering a practical framework for controllable multimodal generation.

Overall, our contributions are as follows: (1) An autoregressive framework for efficient controllable multimodal image generation;  (2) A two-stage training strategy that enables robust alignment and balanced modality integration with substantially reduced computational cost;  (3) Experiments demonstrating the superior efficiency, controllability, and fidelity of \model as a compelling framework for controllable multimodal generation.

\begin{figure*}[t]
\centering
\includegraphics[width=1.0\textwidth]{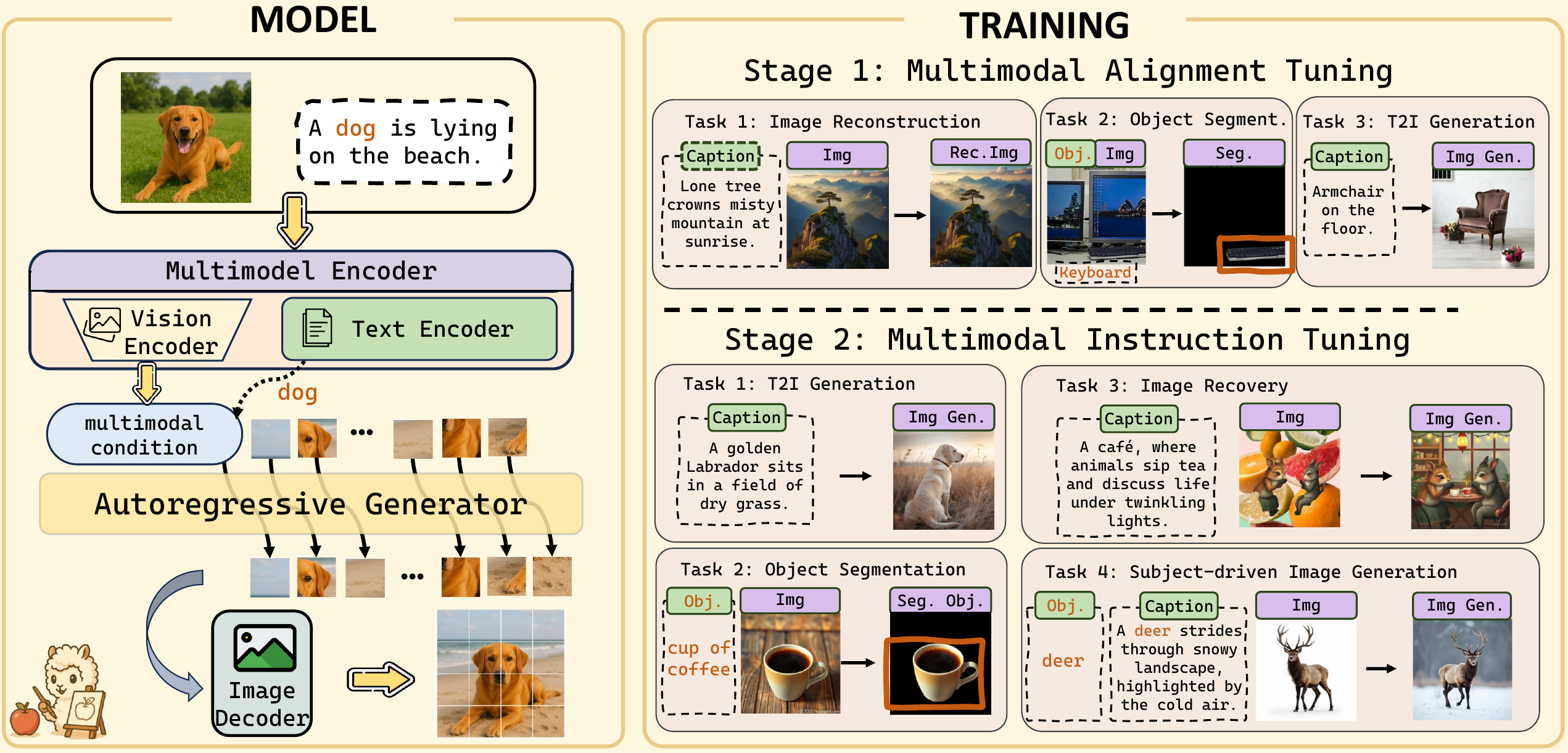}
\caption{Overview of \model. \textbf{Left} panel illustrates model structure, where visual and textual inputs are encoded into a unified latent to guide autoregressive image generation. \textbf{Right} panel highlights two-stage training paradigm: (1) \textbf{Multimodal Alignment Tuning}, enabling pixel and semantic-level alignment between inputs and output tokens; and (2) \textbf{Multimodal Instruction Tuning}, compels model to effectively balance influence of different modalities.}
\label{fig:structure}
\end{figure*} 

\section{Method}
\label{sec:method}

\subsection{Model Design}
\label{sec:model}
As illustrated in \Cref{fig:structure}, \model architecture comprises two core components: a multimodal encoder and an autoregressive generation decoder. 
These components are designed to unify multimodal inputs into a shared embedding and generate image tokens sequentially conditioned on the unified embedding, respectively. 

\textbf{Multimodal Encoder}
The multimodal encoder integrates multimodal inputs from frozen pretrained vision ($\phi_V$) and language ($\phi_L$) encoders into a shared latent space. This module projects visual features from $\phi_V$ into $\phi_L$'s embedding space using a lightweight connector module ($\psi$), yielding a unified multimodal representation $\mathbf{H} = (\mathbf{h}_1, \dots, \mathbf{h}_{M})$, where $\mathbf{h}_j \in \mathbb{R}^{d}$. In practice, we adopt a multi-layer perceptron layer as the connector module that directly projects visual tokens to maintain detail information.

\textbf{Autoregressive Decoder}
A transformer-based autoregressive decoder generates a image token sequence $\mathbf{y} = (y_1, \dots, y_{L})$ conditioned on the prefix $\mathbf{H}$ generated by the multimodal encoder and previously generated tokens $y_{<i}$. It operates in a shared embedding space with the encoder's output and shares the same vocabulary as the VQGAN~\citep{Esser2020TamingTF} that is used for image tokenization.
The generated token sequences are subsequently decoded into images using the VQGAN decoder. This unified autoregressive structure facilitates unified training via next-token prediction.


\subsection{Two-Stage Training Paradigm}
\label{sec:traing_stages}
Effectively aligning different modalities and balancing their influence are crucial challenges for multimodal image generation.
Our two-stage training paradigm directly addresses these issues, moving beyond initial coarse alignment to foster robust understanding and balanced integration of diverse inputs, as illustrated in \Cref{fig:structure}. 

\textbf{Stage 1: Multimodal Alignment Tuning.} 
Although the multimodal encoder can encode multimodal inputs for the generator, we observe that the model tends to interpret visual information semantically like text captions, while neglecting visual details. 
To explicitly strengthen both pixel- and semantic-level alignment and promote richer visual utilization, we introduce three complementary tasks: 
(1) \emph{\textbf{image reconstruction}}, where the model reconstructs an input image (with or without corresponding caption) to enhance pixel-level fidelity; 
(2) \emph{\textbf{object segmentation}}, where the model generates an object-specific segmentation mask given an image and target label, enforcing attention to fine-grained visual details and spatial structures associated with semantic concepts; and 
(3) \emph{\textbf{text-to-image (T2I) generation}}, using image–caption pairs to preserve and reinforce generative capability. 
Notably, combining segmentation with reconstruction prevents trivial copy behavior (i.e., simply replicating input images) by encouraging semantically meaningful and spatially precise outputs. 
This complementary effect is further analyzed in \Cref{abl:stage1}.

\textbf{Stage 2: Multimodal Instruction Tuning.} 
Building upon the alignment and visual fidelity established in Stage~1, this stage aims to equip the model with robust instruction following and cross-modal reasoning abilities. 
To encourage the model to integrate different modalities in a balanced manner, we adopt a multimodal instruction tuning strategy using a curated mixture of tasks. We retain the \emph{text-to-image (T2I) generation} and \emph{object segmentation} tasks from Stage~1, preserving their formulations to maintain foundational skills and stabilize training. In addition, two new tasks are introduced to enhance instruction adherence and balanced modality integration.  
(1) \emph{\textbf{image recovery}} introduces synthetic distortions—such as rotation, resizing, and compositing segmented objects onto random backgrounds—paired with original captions. The model are required to reconstruct the original image from the distorted input and caption, encouraging it to extract essential visual cues and leverage textual information to restore missing details.  
(2) \emph{\textbf{subject-driven image generation}} conditions the model on a reference image, subject label, and textual instruction to generate new images. This task requires preserving the subject’s visual identity while faithfully following textual directives, serving as a comprehensive end-to-end objective.  
Overall, this stage enables the model to integrate visual and textual information in a balanced manner, effectively mitigating over-reliance on a single modality and supporting precise multimodal generation, as further discussed in \Cref{abl:stage2}.

\subsection{Data Construction}
\label{sec:data_construct}

To support our two-stage training paradigm, we construct a multimodal dataset of approximately \textbf{3 million samples}. The dataset integrates open-source resources, synthetic data, and automated annotations to ensure scalability and diversity.  
For \textbf{image reconstruction} and \textbf{T2I generation}, we collect image–text pairs from datasets such as CC12M~\citep{changpinyo2021cc12m} and Midjourney-Niji~\citep{midjourney-niji-1m-llavanext}. To expand domain coverage (e.g., human subjects, artistic scenes), we further synthesize samples using T2I models like Flux.1~\citep{flux}, with prompts generated by advanced LLMs~\citep{gpt4o} to enhance semantic and visual diversity.  
For \textbf{segmentation} and \textbf{image recovery}, which require fine-grained object-level annotations, we design an automated pipeline that combines state-of-the-art LMMs~\citep{Qwen2vl} with segmentation models\citep{sam}. 
For \textbf{subject-driven image generation}, we leverage the OminiControl dataset~\citep{OminiControl}, re-captioned using LMMs to accurately extract subject-relevant descriptions. Additionally, we reverse image pairs to effectively double the usable data.
Data pipeline and formation are detailed in \Cref{sec:data_const}.

\section{Experiments}
\label{sec:experiments}

\subsection{Experimental Setup}
\label{sec:details}


\begin{table*}[t]
\centering
\small
\caption{
Comparison on DreamBench++. Models are ranked by \textbf{CP$\cdotp$PF}, indicating balanced overall multimodal image generation performance. \textbf{CP/PF} ratio reflects overfitting issue toward certain modality. ``*'' denotes model trained \textbf{from scratch}; others are adapted from pre-trained T2I models.
}

\label{tab:main_table}
\resizebox{\textwidth}{!}{%
\begin{tabular}{@{}l@{\hspace{1ex}}c@{\hspace{1ex}}c@{\hspace{1ex}}c@{\hspace{1ex}}c@{\hspace{1ex}}c@{\hspace{1ex}}c@{\hspace{1ex}}c@{\hspace{1ex}}c@{\hspace{1ex}}c@{\hspace{1ex}}c@{\hspace{1ex}}c@{\hspace{1ex}}c@{\hspace{1ex}}c@{\hspace{1ex}}c@{}}
\toprule
\textbf{Method} & \textbf{T2I Model} & \textbf{Train Data} & \textbf{Model Size} 
& \multicolumn{5}{c}{\textbf{Concept Preservation (CP)}} 
& \multicolumn{4}{c}{\textbf{Prompt Following (PF)}} 
& \textbf{CP$\cdotp$PF} & \textbf{CP/PF} \\
\cmidrule(lr){5-9} \cmidrule(lr){10-13}
& & & & \textbf{Animal} & \textbf{Human} & \textbf{Object} & \textbf{Style} & \textbf{Overall} 
& \textbf{Photo.} & \textbf{Style.} & \textbf{Imag.} & \textbf{Overall} & & \\
\midrule
\multicolumn{15}{c}{\textbf{Finetuned on Test Set}} \\
\midrule
Textual Inv.         & SD v1.5      & -     & 860M    & 0.50 & 0.35 & 0.30 & 0.36 & 0.38 & 0.68 & 0.70 & 0.44 & 0.63 & 0.24 & 0.60 \\
DreamBooth           & SD v1.5      & -     & 860M    & 0.64 & 0.19 & 0.49 & 0.48 & 0.49 & 0.79 & 0.78 & 0.51 & 0.72 & 0.36 & 0.68 \\
DreamBooth-L         & SDXL v1.0    & -     & 2.60B   & 0.75 & 0.31 & 0.54 & 0.71 & 0.60 & 0.90 & 0.90 & 0.75 & 0.87 & 0.52 & 0.69 \\

\midrule
\multicolumn{15}{c}{\textbf{Test-Time Tuning-Free Methods}} \\
\midrule
VARGPT-v1.1         & Infinity (VAR)  & 8.3M    & 9.00B  & 0.25 & 0.13 & 0.14 & 0.27 & 0.19 & 0.31 & 0.48 & 0.33 & 0.37 & 0.07 & 0.51 \\
Unified-IO2*         & Unified-IO2  & 8.5B    & 7.00B   & 0.77 & 0.80 & 0.64 & 0.82 & 0.72 & 0.24 & 0.18 & 0.11 & 0.19 & 0.14 & 3.74 \\
Lumina-mGPT          & Chameleon    & 10M     & 7.00B   & 0.95 & 0.97 & 0.89 & 0.85 & 0.91 & 0.31 & 0.25 & 0.15 & 0.25 & 0.23 & 3.63 \\
OmniGen*          &    OmniGen  &   700M   &   3.80B    & 0.39 & 0.35 & 0.25 & 0.22 & 0.30 & 0.70 & 0.71 & 0.67 & 0.70 & 0.21 & 0.43 \\
DreamEngine          & SD3.5        & 21M     & 10.50B  & 0.76 & 0.72 & 0.61 & 0.73 & 0.68 & 0.44 & 0.37 & 0.25 & 0.37 & 0.26 & 1.84 \\

BLIP-Diffusion       & SD v1.5      & 130M    & 1.56B   & 0.67 & 0.56 & 0.47 & 0.51 & 0.55 & 0.58 & 0.52 & 0.30 & 0.50 & 0.27 & 1.11 \\

Kosmos-G             & SD v1.5      & 200M    & 3.00B   & 0.62 & 0.64 & 0.46 & 0.56 & 0.55 & 0.48 & 0.62 & 0.39 & 0.51 & 0.28 & 1.07 \\

IP-A-Plus ViT-H      & SDXL v1.0    & 10M     & 3.00B   & 0.90 & 0.84 & 0.76 & 0.91 & 0.84 & 0.50 & 0.39 & 0.28 & 0.41 & 0.35 & 2.02 \\

Emu2                 & SDXL v1.0    & 16M     & 37.00B  & 0.67 & 0.55 & 0.45 & 0.44 & 0.53 & 0.73 & 0.73 & 0.56 & 0.69 & 0.36 & 0.76 \\

IP-A ViT-G           & SDXL v1.0    & 10M     & 2.50B   & 0.67 & 0.56 & 0.50 & 0.75 & 0.59 & 0.74 & 0.63 & 0.45 & 0.64 & 0.38 & 0.92 \\
\rowcolor{gray!20} \model & LlamaGen & 3M      & 2.31B   & 0.65 & 0.36 & 0.57 & 0.47 & 0.56 & 0.86 & 0.85 & 0.80 & 0.84 & \textbf{0.47} & 0.66 \\
\bottomrule
\end{tabular}%
}
\end{table*}

\textbf{Implementation Details}
We initialized the multimodal encoder with CLIP-Large-Patch14~\citep{Radford2021LearningTV} and FlanT5-XL~\citep{flant5}, which has 224×224 image receptive field, and the generator with LlamaGen-XL~\citep{llamagen}.
Training on 8 A100 GPUs about 1.5 days.
More Details in \Cref{sec:Exp_Details} and \Cref{app:cfg_sweep}.





\textbf{Benchmark \& Metric.} We evaluate \model on \textbf{DreamBench}~\citep{ruiz2023dreamboothfinetuningtexttoimage} and \textbf{DreamBench\texttt{++}}~\citep{peng2025dreambenchpp} benchmarks. 
\textbf{DreamBench} uses CLIP and DINO scores to assess image fidelity and prompt alignment. 
\textbf{DreamBench\texttt{++}} addresses limitations of DreamBench evaluation and evaluate on two axes: \textbf{Concept Preservation (CP)}, measuring the retention of the subject's visual identity, and \textbf{Prompt Following (PF)}, evaluating how accurately the image reflects the text prompt.
A human evaluation study confirms high consistency between GPT-based metrics and human judgments.
Further details are provided in \Cref{app:DreamBench_Plus}.




\textbf{Baselines.}
We compare \model with both fine-tuning and tuning-free methods. 
Fine-tuning baselines include \textbf{Textual Inversion}~\citep{gal2022imageworthwordpersonalizing} and \textbf{DreamBooth}~\citep{peng2025dreambenchpp}.
Tuning-free baselines include diffusion-based models—\textbf{BLIP-Diffusion}~\citep{li2023blipdiffusionpretrainedsubjectrepresentation}, \textbf{Emu2}~\citep{emu2}, \textbf{IP-Adapter}~\citep{ye2023ip-adapter}, \textbf{OmniGen}~\citep{OmniGen}, and \textbf{DreamEngine}~\citep{dreamengine}—and AR models such as \textbf{Unified-IO 2}~\citep{lu2023unifiedio2}, \textbf{Lumina-mGPT}~\citep{2024lumina}, and \textbf{VARGPT-v1.1}~\citep{zhuang2025vargptv11improvevisualautoregressive}.

\subsection{Main Results}
\label{sec:main}
\Cref{tab:main_table} comprehensively evaluates our proposed autoregressive (AR) framework on the DreamBench++ benchmark, comparing it with diffusion-based and autoregressive-based baselines.
\model demonstrates highly competitive performance, particularly in achieving a strong balance between the guidance of both input modalities. Notably, this is achieved despite utilizing significantly fewer training resources and suboptimal model components compared to the state-of-the-art baselines.

\textbf{Overall Performance.}
\model achieves a strong balance between concept fidelity and prompt alignment, resulting in the high \textbf{CP$\cdotp$PF} score. Quantitative examples can be found in \Cref{fig:dream_exp}.
It rivals fine-tuned methods such as DreamBooth-LoRA while significantly outperforming test-time tuning-free baselines including OmniGen and DreamEngine.  
A key strength of \model lies in its ability to harmoniously integrate multimodal inputs. Several strong baselines, including autoregressive models like Lumina-mGPT and Unified-IO2, achieve high Concept Preservation (CP) but extremely low Prompt Following (PF), leading to high \textbf{CP/PF} ratios and indicating over-reliance on visual references while neglecting textual guidance. Even for models like VARGPT-v1.1, which has the advanced VAR generator, despite producing visually appealing results, fail to consistently follow multimodal instructions.    
In contrast, \model delivers low CP/PF scores compared to most baselines, demonstrating effective and controlled integration of both visual and textual guidance.

\textbf{Training Efficiency.}
A notable advantage of \model lies in its training efficiency. It is trained on only 3 million image-text pairs across two stages, substantially less than leading baselines, such as Emu2 (16M), Kosmos-G (200M), and DreamEngine (21M).
Beyond the reduced data requirements, the training process is highly resource-efficient: the entire training process completes in 1.5 days with 8 GPUs. This contrasts sharply with other baselines, such as Kosmos-G, which necessitates 256 GPUs over three days.
Despite this dramatically reduced computational and data budgets, \model achieves competitive performance with balanced performance, highlighting its efficiency and effectiveness.
Furthermore, \model remains highly competitive in size compared to larger counterparts, highlighting our effectiveness.


%

\textbf{Discussion and Connection to Methodology.}
The strong performance of \model—particularly its balanced multimodal generation and training efficiency—stems from its \textbf{autoregressive nature} and \textbf{two-stage training paradigm}. 
The \textbf{autoregressive design}, which generates image tokens sequentially conditioned on a unified multimodal prefix, enables fine-grained, token-level alignment between multimodal inputs and outputs. 
This direct alignment improves controllability and ensures generated results accurately follow both text and visual guidance. 
Notably, \textbf{two-stage training paradigm} promotes balanced multimodal control, mitigating the dominance of any single modality and yielding significantly higher CP$\cdot$PF scores compared to baselines.  
Notably, \model achieves strong results despite using \textbf{relatively suboptimal components}.
While baselines rely on advanced models such as Qwen-2.5 and SD3, our implementation employs Flan-T5 as the encoder and LlamaGen as the generator—both substantially weaker than their counterparts, as shown in \Cref{tab:exp-geneval}.  
This highlights that the improvements are stem from methodological design rather than model or data scale.

\begin{table*}[t]
\centering
\small
\caption{Ablation results on DreamBench++ and DreamBench.
}
\label{tab:ablation_unified}
\resizebox{\linewidth}{!}{%
\begin{tabular}{lcccccc}
\toprule
\multirow{2}{*}{\textbf{Method}} &
\multicolumn{3}{c}{\textbf{DreamBench++}} &
\multicolumn{3}{c}{\textbf{DreamBench}} \\
\cmidrule(lr){2-4} \cmidrule(lr){5-7}
 & \textbf{CP} & \textbf{PF} & \textbf{CP$\cdotp$PF} 
 & \textbf{DINOv1} & \textbf{CLIP-I} & \textbf{CLIP-T} \\
\midrule
\textit{w/o Obj. Seg. in Stage 1}   
& $0.252 \pm 0.004$ & $0.479 \pm 0.005$ & $0.121$ 
& $56.113 \pm 0.082$  & $74.384 \pm 0.071$  & $23.965 \pm 0.038$ \\

\textit{w/o Stage 1 Alignment}     
& $0.179 \pm 0.002$ & $0.673 \pm 0.012$ & $0.120$ 
& $33.523 \pm 0.111$  & $67.705 \pm 0.101$  & $28.263 \pm 0.155$ \\

\textit{w/o Image Recovery}        
& $0.661 \pm 0.007$ & $0.284 \pm 0.004$ & $0.188$ 
& $74.471 \pm 0.321$ & $81.280 \pm 0.094$ & $24.210 \pm 0.022$ \\

\textit{w/o Object Segmentation}   
& $0.412 \pm 0.002$ & $0.918 \pm 0.003$ & $0.378$ 
& $57.221 \pm 0.119$  & $76.269 \pm 0.084$  & $31.078 \pm 0.050$ \\

\textit{w/o Multimodal T2I Task}   
& $0.407 \pm 0.004$ & $0.910 \pm 0.004$ & $0.370$ 
& $58.880 \pm 0.143$  & $76.529 \pm 0.102$  & $30.483 \pm 0.002$ \\

\rowcolor{gray!10}
\textbf{\model}         
& $0.555 \pm 0.006$ & $0.839 \pm 0.002$ & $\mathbf{0.466}$ 
& $70.853 \pm 0.327$  & $80.911 \pm 0.053$  & $29.071 \pm 0.080$ \\
\bottomrule
\end{tabular}%
}
\end{table*}

\subsection{Ablation Study}
\label{sec:ablation}

We conduct an ablation study on DreamBench++ and DreamBench focusing on two central questions: (1) How critical is Stage 1 for robust multimodal alignment? and (2) What role does each training task play in shaping model’s multimodal generation behavior?
Following prior work~\citep{peng2025dreambenchpp}, we report \textbf{CP$\cdot$PF} as a primary measure of multimodal image generation ability.

\textbf{Importance of Stage 1: Foundational Multimodal Alignment.} \label{abl:stage1}
As shown in \Cref{tab:ablation_unified}, removing Stage 1 leads to severe performance drop, underscoring its foundational role. On DreamBench++, the CP score drops from 0.555 to 0.179, indicating a major loss in visual identity preservation, with PF also significantly reduced. Similar trends appear on DreamBench.
Ablating only \textbf{object segmentation task} in Stage 1 (\textit{w/o Obj. Seg. in Stage 1}) also hampers model performance.
While the remaining image reconstruction task supports pixel-level alignment, allowing for reconstruction of input images, it inadvertently leads the model to exhibit a copy-paste behavior, failing to capture semantic and visual information of input images.
It confirms that reconstruction alone is insufficient for robust multimodal alignment.
Overall, these results highlight that without Stage 1, the model struggles to ground visual concepts from images, severely impairing its visual preservation ability.

\begin{table*}[t]
\centering
\caption{Controllable experiments between \model and Kosmos-G in DreamBench++ benchmark.}
\label{tab:comparison}
\resizebox{0.9\textwidth}{!}{%
\begin{tabular}{@{}lcccccccccc@{}}
\toprule
\textbf{Method} & \multicolumn{5}{c}{\textbf{Concept Preservation (CP)}} & \multicolumn{4}{c}{\textbf{Prompt Following (PF)}} & \textbf{CP$\cdotp$PF} \\ 
\cmidrule(lr){2-6} \cmidrule(lr){7-10}
 & \textbf{Animal} & \textbf{Human} & \textbf{Object} & \textbf{Style} & \textbf{Overall} & \textbf{Photo.} & \textbf{Style.} & \textbf{Imag.} & \textbf{Overall} & \\ 
\midrule
Kosmos-G  & 0.17 & 0.08 & 0.14 & 0.18 & 0.15 & 0.72 & 0.71 & 0.68 & 0.71 & 0.11 \\
\model     & 0.65 & 0.36 & 0.57 & 0.47 & \textbf{0.55} & 0.86 & 0.85 & 0.80 &\textbf{0.84} & \textbf{0.47} \\ 
\bottomrule
\end{tabular}%
}
\end{table*}

\textbf{Contributions of Different Training tasks in Stage 2.} \label{abl:stage2} 
The distinct contributions of different Stage 2 tasks are also evident in \Cref{tab:ablation_unified}.
Excluding the \textit{Image Recovery} task leads to a sharp imbalance: while visual preservation metrics (CP, DINOv1, CLIP-I) show a notable increase, instruction following ability (PF and CLIP-T score) critically drops, showing an overfitting to visual features. This underscores that image recovery acts as a critical regularizer, encouraging the model to reconstruct incomplete visual contexts guided by text prompt, thereby fostering a balanced use of different modalities.
Conversely, ablating either the \textit{Object Segmentation} or the \textit{Subject-Driven Image generation} significantly degrades visual preservation ability, as these tasks prompt the model to utilize the visual features of the input image effectively to generate images.
These results demonstrate that image recovery ensures cross-modal balance, while object segmentation and subject-driven generation enhance the model's ability to extract and utilize detailed visual information for image generation.
Additional task weight sweeping experiments in \Cref{app:task_weight} further confirm the CP$\cdotp$PF trade-off controlled by the task mixture.


\subsection{Analysis}
\label{sec:Analysis}

\textbf{Efficiency and Effectiveness: AR vs. Diffusion.}
To evaluate the efficiency of our AR framework against diffusion-based methods, we conducted a controlled comparison with \textbf{Kosmos-G}~\citep{Kosmos-G}, a representative LMM-augmented diffusion model.Both models were trained from \textbf{similar initializations} on \textbf{same training data} to ensure a fair comparison. Shown in \Cref{tab:comparison}, despite Kosmos-G employing a superior SD1.5 generator and Kosmos-1 encoder, \model, still shown significantly better performance with limited training data, highlighting the effectiveness of our framework.

\begin{table}[htbp] 
    \centering
    \small
    \footnotesize 
    \caption{Image reconstruction performance.}
    \label{tab:reconstruct_l2}
    \resizebox{0.9\linewidth}{!}{
    \begin{tabular}{@{}lcc@{}}
    \toprule
    \textbf{Method} & \textbf{COCO ($\downarrow$)} & \textbf{JourneyDB ($\downarrow$)} \\
    \midrule
    SeedTokenizer   & 0.5102 & 0.5291 \\
    SEED-X          & 0.4317 & 0.4352 \\
    EMU2-Gen        & 0.3828 & 0.2869 \\
    DreamEngine     & \underline{0.2065} & \underline{0.2052} \\
    \midrule
    \textbf{\model} & \textbf{0.1008} & \textbf{0.0867} \\
    \bottomrule
    \end{tabular}
    }
\end{table}

\textbf{Comparison of Architecture Variants. }
We compare architectural variants of \model that differ in how visual features are connected to the generator.
Since the multimodal encoder produces hundreds of visual tokens per image, the inflated context length poses significant computational challenges, particularly in multi-image settings. Token compression alleviates this issue but may compromise fine-grained visual fidelity.
To navigate this trade-off, as shown in \Cref{tab:ablation_unified}, we evaluate a query-based connector that leverages a Query-Former~\cite{li2023blip2} to compress long visual sequences into fixed-size representations, guided by textual queries~\citep{li2023blipdiffusionpretrainedsubjectrepresentation} highlighting important concepts in images.
The results reveals that while the query-based approach substantially reduces computational cost, it struggles to preserve fine-grained visual details crucial for generative fidelity, even guided by textual queries.
Consequently, the MLP-based connector(\model) achieves higher fidelity, especially for human and object regions, whereas the query-based variant offers a favorable trade-off between efficiency and performance.
Despite these differences, both variants exhibit competitive performance compared to other baselines. 

\begin{table*}[t]
\centering
\small
\caption{Analysis on architecture design, multi-image training and reinforcement learning}
\label{tab:ablation_combined}
\resizebox{\linewidth}{!}{%
\begin{tabular}{lcccccc}
\toprule
\multirow{2}{*}{\textbf{Method}} &
\multicolumn{3}{c}{\textbf{DreamBench++}} &
\multicolumn{3}{c}{\textbf{DreamBench}} \\
\cmidrule(lr){2-4} \cmidrule(lr){5-7}
 & \textbf{CP} & \textbf{PF} & \textbf{CP$\cdotp$PF} 
 & \textbf{DINOv1} & \textbf{CLIP-I} & \textbf{CLIP-T} \\
\midrule

\rowcolor{gray!10}
\textbf{\model}   
& $0.555 \pm 0.006$ & $0.839 \pm 0.002$ & $0.466$ 
& $70.853 \pm 0.327$  & $80.911 \pm 0.053$  & $29.071 \pm 0.080$ \\

\textit{w. Query-Variants}        
& $0.421 \pm 0.002$ & $0.882 \pm 0.000$ & $0.371$ 
& $54.518 \pm 0.317$  & $76.306 \pm 0.114$  & $30.792 \pm 0.040$ \\

\textit{w. Multi-image}         
& $0.586 \pm 0.006$ & $0.829 \pm 0.005$ & $0.486$ 
& $72.487 \pm 0.147$ & $81.857 \pm 0.152$ & $28.545 \pm 0.043$ \\
\textit{w. GRPO}         
& $0.609 \pm 0.006$ & $0.866 \pm 0.004$ & $\mathbf{0.527}$ 
& $77.176 \pm 0.199$ & $83.343 \pm 0.287$ & $29.810 \pm 0.059$ \\
\bottomrule
\end{tabular}%
}
\end{table*}

\textbf{Effect of Multi-Image Training.}
To assess the benefits of richer visual context, we further trained the model using a mix of Stage 2 data and additional multi-subject task(reconstructing images based on segmented objects and image caption) generated via our data construction pipeline. As shown in \Cref{tab:ablation_combined}, \textit{w. MultiImage Training} achieves a higher CP$\cdotp$PF score (0.49), improving CP to 0.60 while maintaining a strong PF score.
This emphasizes the advantage of enhanced visual context in training, prompting the model to efficiently handle and integrate information from multiple visual inputs, thereby improving its ability to preserve visual details in complex multimodal scenarios.


\begin{figure}[htbp] 
    \centering
    \includegraphics[width=\linewidth]{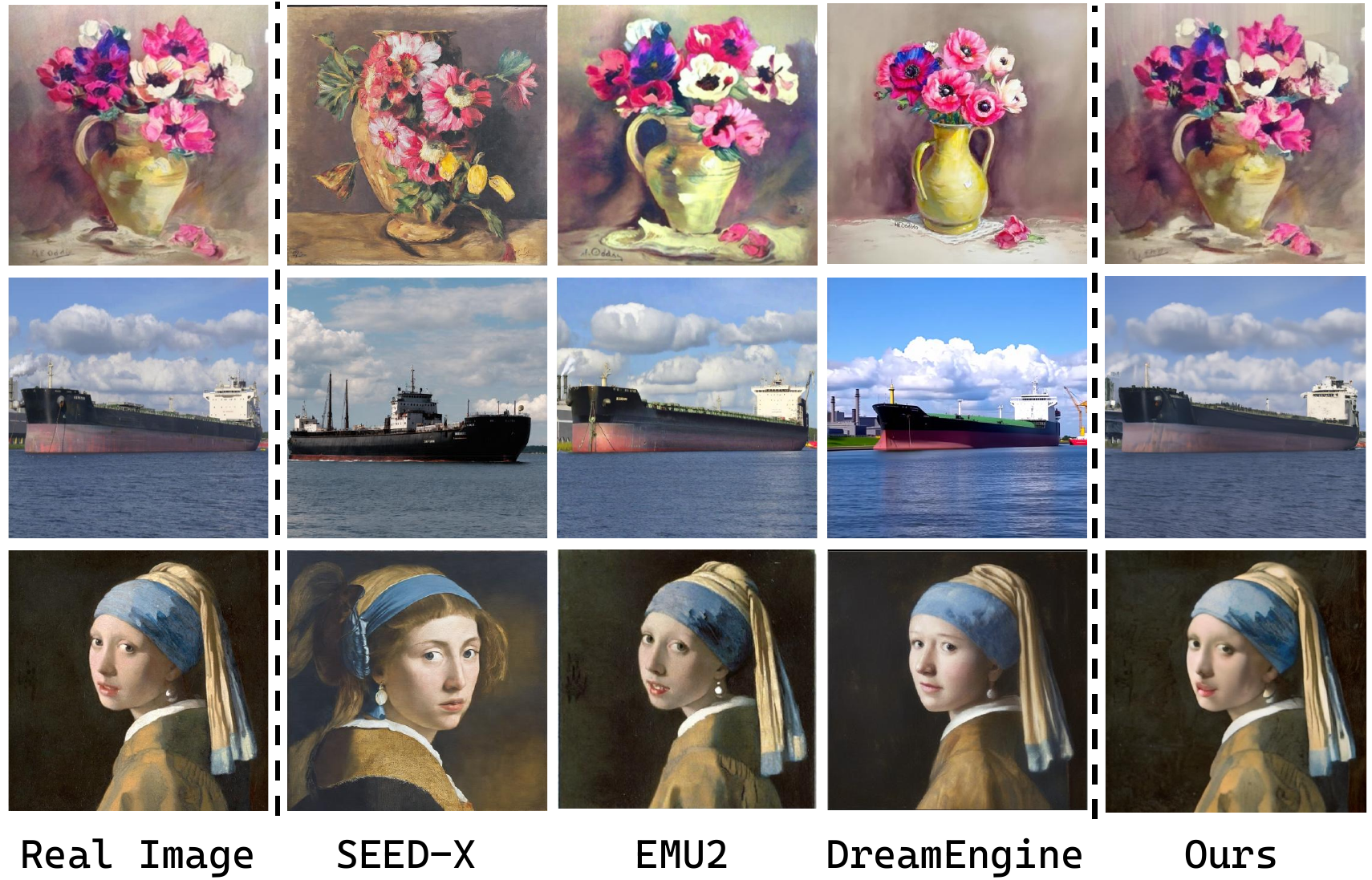}
    \caption{Qualitative study on Image Reconstruction.}
    \label{fig:reconstruction}
\end{figure}

\textbf{Image Reconstruction Fidelity.}
To quantify visual detail preservation in our framework, we evaluate \model on the Image Reconstruction Benchmark~\citep{dreamengine}, which measures similarity between input and reconstructed images. After fine‑tuning on reconstruction task for 1{,}000 steps, we compare the generated outputs with their originals, following previous work~\citep{dreamengine}. As shown in \Cref{tab:reconstruct_l2}, \model outperforms strong baselines with comparable architectures—SeedTokenizer~\citep{seed-tokenizer}, EMU2~\citep{emu2}, SeedX~\citep{2024SeedX}, and DreamEngine~\citep{dreamengine}—all of which couple LMMs with diffusion backbones.
\model achieves best reconstruction quality even with a $224{\times}224$ receptive field, while others varied from 384x384 to 512x512.


\textbf{Versatility Across Different Multimodal Tasks.}
To explore broader applicabilities of our framework, we evaluate its adaptability across diverse tasks, including image segmentation,  multi-image generation and multimodal in-context image generation. This was achieved with brief fine-tuning on relevant datasets, as detailed in~\Cref{app:applications}. Qualitative results in \Cref{fig:examples}, \Cref{fig:segmentation_example}, \Cref{fig:multi_image} and \Cref{fig:icl_example} show that \model produces coherent outputs that adhere to provided constraints without any architectural modifications.
While achieving performance in each specific domain would necessitate more specialized training and potentially more powerful components, these initial results underscore our framework's versatility and its potential as an effective foundation for multimodal image generation.
 
\textbf{Effect of Reinforcement Learning on Multimodal Generation.}
Given the autoregressive nature of our framework, \model naturally supports token-level reinforcement learning (RL), enabling direct optimization of generation behaviors. 
We apply the GRPO~\citep{grpo} following the AR-GRPO~\citep{yuan2025argrpotrainingautoregressiveimage} to explore potential of RL in enhancing multimodal controllable generation. 
Specifically, the model is fine-tuned with reward signals from combined reward models, encouraging outputs that better align with multimodal guidance while preserving visual coherence. 
As shown in \Cref{tab:ablation_combined}, \textit{w.GRPO} achieves the highest CP$\cdotp$PF score (0.527), surpassing other variants. 
More details can be found in \Cref{app:rl}.

\section{Related Work}
\label{sec:relatedwork}

\subsection{Image Generation with Complex Multimodal Control}
Recent advances in diffusion models enable image generation conditioned on multimodal inputs such as canny edges~\citep{controlnet} or reference images~\citep{ultraEdit,SDEdit}. 
DreamBooth~\citep{ruiz2023dreamboothfinetuningtexttoimage} enables subject-specific fine-tuning but limits generalization. SuTI~\citep{suti} addresses it with scalable data and training.
To enhance flexibility, recent work integrates LMMs with diffusion models~\citep{koh2023GILL,sun2023emu1,dreamllm,unimo}. Approaches like Kosmos-G~\citep{Kosmos-G}, Emu-2~\citep{emu2}, Seed-X~\citep{2024SeedX}, and DreamEngine~\citep{dreamengine} explore more complex multimodal prompt and fine-grained multimodal control. Yet, balancing guidance from diverse modalities remains a core challenge~\citep{han2024emmatexttoimagediffusionmodel,ye2023ip-adapter,RealCustom++}. 
EMMA~\citep{han2024emmatexttoimagediffusionmodel} employs a gated perceiver resampler for dynamic signal integration, while RealCustom++\citep{RealCustom++} disentangles subject identity and textual fidelity via cross-layer projectors. OmniControl~\citep{OminiControl} introduces a bias term into multimodal attention. 
Recently, large-scale systems such as Qwen-Image~\citep{wu2025qwenimagetechnicalreport}, Begal~\citep{begal}, and Nano-Banana~\citep{comanici2025gemini25pushingfrontier} further demonstrate impressive multimodal generation capabilities through large-scale training on extensive multimodal corpora across thousands of GPUs.
Nonetheless, these methods still demand substantial resources, and achieving efficient multimodal integration remains an open problem.

\subsection{Autoregressive Multimodal Generation}
Autoregressive models have driven progress in T2I generation, from DALL·E~\cite{DALLE} to LlamaGen~\citep{llamagen} and GPT4O~\citep{gpt4o}.
Recent work extends it to multimodal settings: Models like Chameleon~\citep{chameleonteam2024chameleon}, LWM~\cite{LWM}, AnyGPT~\citep{Zhan2024AnyGPT}, and EMU3~\citep{Emu3} treat text and images as unified token sequences via early-fusion transformers, yet still emphasize text-to-image generation with limited support for multimodal conditioning.
Janus~\citep{Janus} decouples visual understanding and generation via distinct pathways but lacks support for multimodal image generation. 
MUSE-VL~\cite{musevl} and VILA-U~\citep{VILA-U} align discrete visual tokens with text to improve perception, but remain oriented toward understanding tasks rather than image generation.
Unified-IO2~\citep{lu2023unifiedio2} is trained autoregressively from scratch for both understanding and generation across modalities, while Lumina-mGPT~\cite{2024lumina} and VARGPT~\citep{zhuang2025vargptv11improvevisualautoregressive} enhances Chameleon and VAR~\citep{han2024infinityscalingbitwiseautoregressive} with supervised fine-tuning for broader multimodal tasks. 
Overall, while models like VILA-U~\cite{VILA-U}, EMU3~\cite{Emu3}, and Janus~\cite{Janus} have advanced text-to-image generation, robust multimodal conditional generation~\citep{wu2025nepautoregressiveimageediting,cai2025mmgrmultimodalgenerativereasoning} remains an open and underexplored challenge.

\section{Conclusion}
\label{sec:conclusion}

In this work, we introduced a controllable and efficient autoregressive framework for complex multimodal image generation, offering a compelling alternative to diffusion-based methods.
By unifying multimodal inputs within an AR model and leveraging a two-stage training paradigm, our method achieves state-of-the-art performance on challenging benchmarks—despite a modest model size, suboptimal base component, and limited training resources.
These results underscore the efficiency, scalability, and controllability of our method, establishing it as an efficient foundation for building versatile, fine-grained visual generation systems capable of handling complex multimodal prompts.

\section*{Limitations}
\label{app:limitation}

Our work introduces a resource-efficient framework for multimodal-conditioned image generation, primarily aimed at exploring how to {balance multimodal guidance for controllable generation rather than maximizing absolute generation quality. 
Accordingly, our evaluation focuses on verifying the effectiveness of the proposed training paradigm and architectural design under limited-resource settings, using models of comparable capacity and scale.
However, the current performance of our approach is inherently constrained by the limitations of generative backbone models. 
Unlike recent unified or large-scale models trained with thousands of GPUs or on extensive datasets, our experiments are conducted on a smaller scale using publicly available resources. 
Consequently, \model exhibits shortcomings in text-to-image generation, including spatial reasoning, object counting, fine-grained human rendering, and stylization. These limitations reflect the current gap between current SOTA diffusion and autoregressive architectures in terms of generation fidelity and domain generalization.
Future work can readily extend this framework by replacing backbone modules and scaling data resources to further enhance image quality.
Additionally, while our training data is sourced from publicly available datasets and our synthetic data pipeline includes NSFW safeguards, a comprehensive evaluation of safety, fairness, and potential misuse remains lacking. Future work should incorporate thorough assessments of model biases and unintended behaviors.
Finally, while our framework demonstrates strong versatility across diverse multimodal tasks, achieving competitive performance in specific domains may require more specialized training and the integration of more powerful multimodal encoders and generators.

\section*{Acknowledgments}
This research was supported in part by the cloud credits from the NVIDIA Academic Grant Program. The views and conclusions expressed in this work are those of the authors and should not be interpreted as representing the official policies or endorsements of NVIDIA.

\bibliography{custom}

\appendix

\clearpage
\newpage
\appendix
\section{Appendix}
\noindent This Appendix is organized as follows.

\begin{itemize}[left=0pt]
    \item In \Cref{sec:preliminary}, we describe the autoregressive training objective for our framework.
    \item In \Cref{sec:Exp_Details}, we provide the training details of \model, including initialization (\Cref{sec:init_Details}), training procedures (\Cref{sec:tra_Details}), multi-image training strategy (\Cref{sub:multi_init_Details}), and RL training with GRPO (\Cref{app:rl}).
    \item In \Cref{sup:t2i}, we show quantitative evaluations of our method on text-to-image generation benchmarks.
    \item In \Cref{sec:data_const}, we detail our data construction pipeline and the dataset details used across the two-stage training.
    \item In \Cref{sec:Experiment_Details}, we elaborate on the experimental setup, including datasets and metrics (\Cref{sec:Benchmark_Details}), as well as detailed descriptions of baseline methods (\Cref{sec:Baselines_Details}).
    \item In \Cref{sec:Qualitative_Study}, we present qualitative results that demonstrate the capabilities of \model in various settings, such as image reconstruction (\Cref{sec:Reconstruction}), segmentation (\Cref{sec:Segmentation}), multi-image generation (\Cref{sec:Multi-Image}), and in-context image generation (\Cref{sec:mmicl}).
    \item In \Cref{app:applications}, we demonstrate the versatility of \model across diverse multimodal generation tasks, including segmentation, subject-driven generation, and multimodal in-context learning.
    \item In \Cref{app:human_eval}, we present the human evaluation study validating the reliability of GPT-based metrics.
    \item In \Cref{app:cp_pf_tradeoff}, we analyze CP--PF trade-offs via task weight and CFG sweeps.
\end{itemize}

\subsection{Preliminary}
\label{sec:preliminary}
\textbf{Training Objective}
Our model employs \emph{teacher forcing} to predict image tokens, conditioned on (i) previously generated tokens and (ii) multimodal context $\mathbf{h}$. 
Given the multimodal condition: $\mathbf{c}^{(0)} = \{\mathcal{I}, \mathcal{T}\}$ (visual and textual inputs),
a multimodal encoder~\(\phi\) first encodes $\mathbf{c}^{(0)}$ and subsequently uses an MLP layer to project them into the space of the image decoder to form a unified representation $\mathbf{h}$:
\begin{equation}
\small
    \mathbf{H} = \text{MLP}(\phi\bigl(\mathbf{c}^{(0)}\bigr))
    = (\mathbf{h}_1, \dots, \mathbf{h}_{M}) \in \mathbb{R}^{M\times d}, \space \mathbf{h}_j \in \mathbb{R}^{d}.
    \label{eq:enc}
\end{equation}
where $M$ is the number of conditioning tokens, and $d$ is the dimension of the latent embeddings.
Then, the AR decoder $\theta$, conditioned on $\mathbf{h}$, generates image sequence $\mathbf{y} = (y_1, \dots, y_L)$ as follows:
\begin{equation}
    \small
    \theta(\mathbf{y} \mid \mathbf{H}) = \prod_{i=1}^{L} \theta\bigl(y_i \mid y_{<i}, \mathbf{H}\bigr).
    \label{eq:factor}
\end{equation}
The training objective is to minimize the token-level cross-entropy loss by \emph{teacher forcing} on data $\mathcal{D}$:
\begin{equation}
\small
    \mathcal{L}_{\text{CE}}(\theta, \phi)
    = -\mathbb{E}_{(\mathbf{y}, \mathbf{c}^{(0)}) \sim \mathcal{D}}
    \left[
        \sum_{i=1}^{L}
        \log \theta\bigl(y_i \mid y_{<i}, \mathbf{H} \bigr)
    \right].
    \label{eq:ce}
\end{equation}

\textbf{Classifier-free Guidance} 
To enhance multimodal generation controllability, we apply Classifier-Free Guidance (CFG)~\citep{llamagen}. During training, multimodal conditioning \(\mathbf{H}\) is replaced by a learned unconditional embedding \(\mathbf{H}_u\) with probability \(p\)~\citep{zhao-etal-2025-looking}. At inference time, token logits \(\ell_g\) are recalculated by interpolating between the conditional logits \(\ell_c\) (from \(\mathbf{H}\)) and unconditional logits \(\ell_u\) (from \(\mathbf{H}_u\)), controlled by a scaling parameter \(\lambda\):
\(\ell_g = \ell_u + (\ell_c - \ell_u) \times \lambda\).

\section{Training Details}
\label{sec:Exp_Details}

\subsection{Initialization Details}
\label{sec:init_Details}

The multimodal encoder is initialized using the vision encoder from CLIP-Large-Patch14~\citep{Radford2021LearningTV}, with an image receptive field of $224 \times 224$, and the FlanT5-XL encoder~\citep{flant5}, with a context length of 512 tokens. This encoder converts each image into 256 tokens for use as context in the generator.

To implement the MLP-based projection, we train the MLP projector on the LLaVA-CC3M-Pretrain-595K dataset~\citep{liu2023llava}, following the alignment training setup used by LLaVA. Specifically, we freeze both the vision and text encoders (CLIP-Large-Patch14 and FlanT5-XL, respectively) and train only the MLP layers. The resulting pretrained MLP layers are then directly incorporated into the multimodal encoder of \model.

\begin{figure*}[htbp]
\centering
\includegraphics[width=0.8\textwidth]{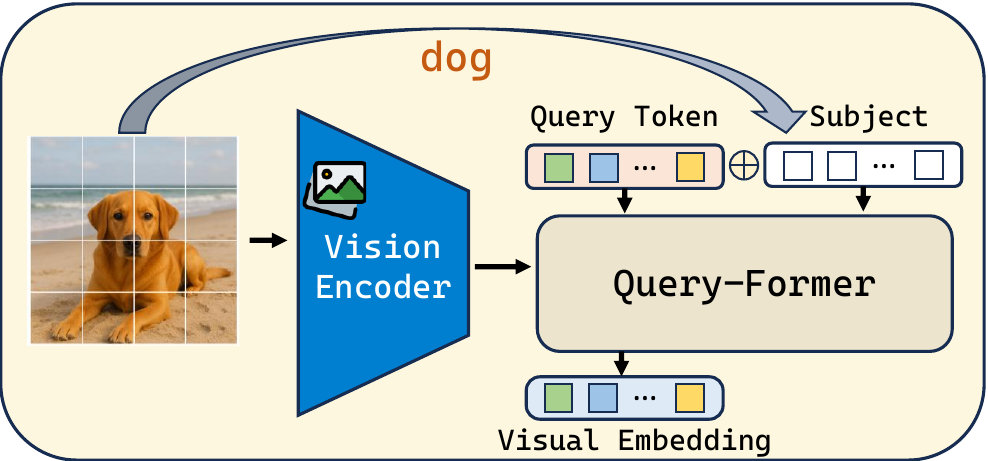}
\caption{Overview of text-guided visual distillation using the Query-based variant of \model.}
\label{fig:qformer}
\end{figure*}

The projector consists of a two-layer MLP with an intermediate dimension of 4,096, employing SiLU activation functions.
The autoregressive generator is initialized from LlamaGen-XL~\citep{llamagen} with 775 million parameters. 
However, the original LlamaGen implementation contains a fundamental error in its 2D Rotary Positional Embedding~\citep{lu2023unifiedio2,EVA-02} (ROPE) mechanism\footnote{\url{https://github.com/FoundationVision/LlamaGen/issues/54}}, which leads to a loss of information in the query and key vectors during attention computation. To address this, we correct the ROPE implementation in our code and continue training the revised model on both the Midjourney dataset~\citep{midjourney-niji-1m-llavanext} and the LAION-COCO dataset used in LlamaGen pretraining, effectively replicating the original pretraining conditions. This continued training enables the model to adapt to the corrected ROPE mechanism. The resulting model is then used to initialize our autoregressive generator.

\subsection{Training Procedure}
\label{sec:tra_Details}

The model training comprises two distinct stages:

\textbf{Stage 1}: We freeze the multimodal encoder and train only the projector and generator modules for one epoch, using a global batch size of 128. Optimization employs the Adam optimizer with an initial learning rate of $5 \times 10^{-4}$, a linear warm-up over the initial 5

\textbf{Stage 2}: We fine-tune the entire model, excluding the vision encoder, for two epochs. The learning rate is reduced to $1 \times 10^{-4}$, with all other optimization settings remaining consistent with Stage 1. This phase primarily enhances cross-modal interactions and improves conditional image generation capabilities from combined visual and textual inputs.

Training is conducted across 8 NVIDIA A100 GPUs, each equipped with 80 GB memory, taking approximately 1.5 days. Specifically, Stage 1 training involves 2.48 million data points over a single epoch, completed in roughly 14 hours. Stage 2 training utilizes 1.3 million data points over two epochs, taking approximately 20 hours in total.

Ablation studies follow the same training schedule, with one epoch of training on Stage 1 data, followed by two epochs on Stage 2 data.

\begin{figure*}[htbp]
    \centering
    \includegraphics[width=\linewidth]{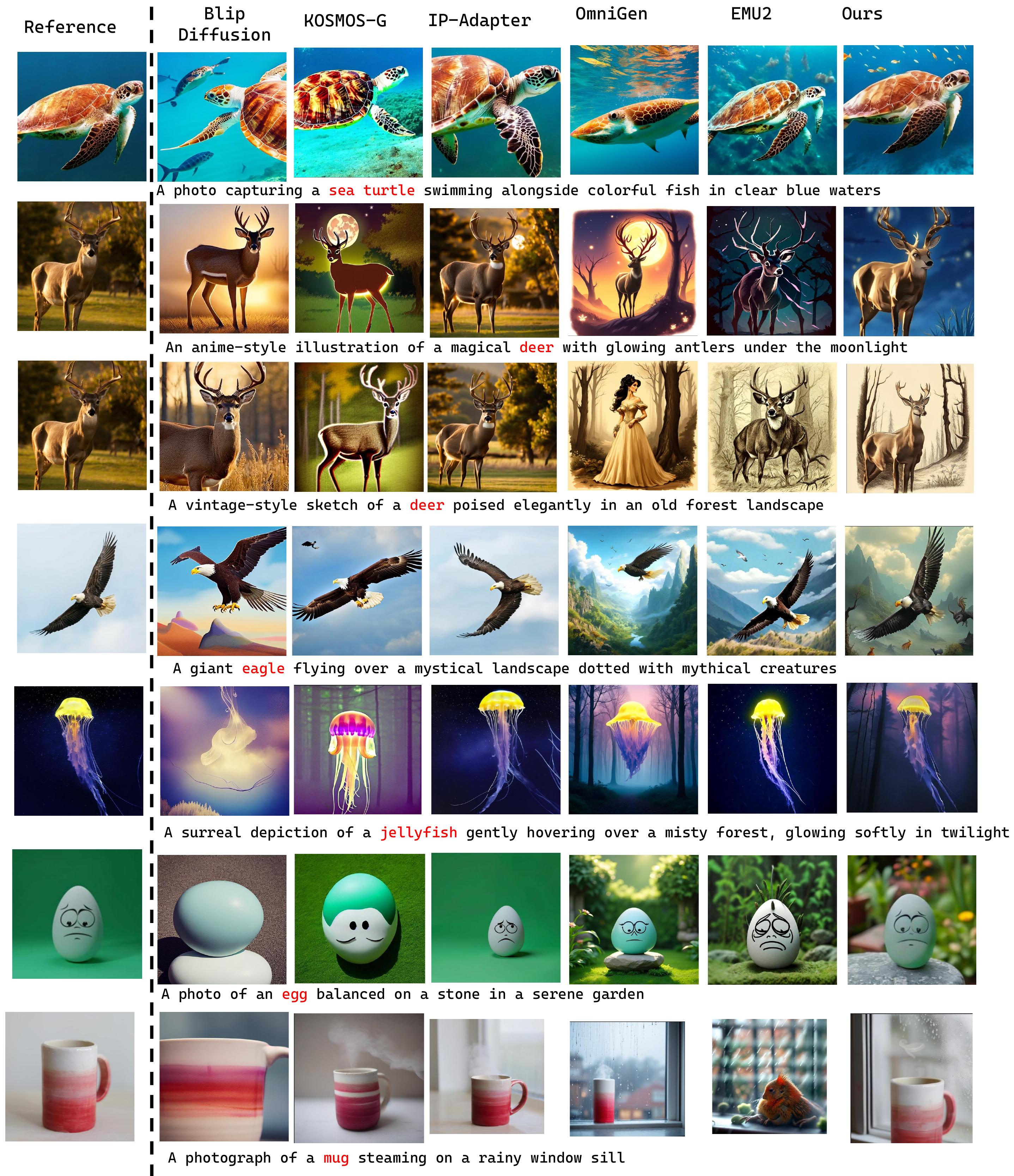}
    \caption{Qualitative examples of different methods compared to \model on DreamBench++.}
    \label{fig:dream_exp}
\end{figure*}

\subsection{Multi-Image Training}
\label{sub:multi_init_Details}
In the \textbf{multi-image training} scenario, the context length of \model is expanded to 1,280 tokens to accommodate up to 4 images per context. For the Query-based variant of \model, token compression techniques enable processing up to 14 images per context with 512 context length.

We utilize 1.5 million multi-image samples, each comprising segmented sub-images accompanied by textual descriptions. The model is trained to reconstruct original images based on these segmented inputs and their corresponding captions. Training incorporates a mixture of Stage 2 data and multi-image samples for an additional epoch.

Qualitative assessments, presented in \Cref{fig:multi}, demonstrate that multi-image training significantly enhances the model's capability to preserve detailed visual information in complex multimodal contexts.

\subsection{GRPO Fine-Tuning for Autoregressive Multimodal Generation}
\label{app:rl}

To explore reinforcement learning~\citep{si2025goal,si2026teaching} for controllable multimodal generation, we adopt the Group Relative Policy Optimization (GRPO) algorithm~\citep{grpo} under the autoregressive (AR) training paradigm, following the AR-GRPO setup~\citep{yuan2025argrpotrainingautoregressiveimage}. Given the autoregressive nature of our framework, this enables token-level optimization of generation behaviors, aligning output quality directly with multimodal guidance.

We apply the vanilla GRPO implementation from the \texttt{trl}~\citep{TRL} library. The model is fine-tuned for 600 steps with a batch size of 12 and a group size of 8, resulting in 96 sampled generations per update. The training corpus includes 73K multimodal samples from UNO-1M~\citep{uno} and 60K text-to-image samples from BLIP-3O~\citep{chen2025blip3ofamilyfullyopen}. KL regularization is applied toward a frozen reference policy to maintain training stability.

We employ a composite reward that integrates multiple complementary signals reflecting perceptual quality, aesthetic appeal, and semantic fidelity. Specifically, for text-to-image (T2I) generation, the reward combines \textit{HPSv2}, \textit{Aesthetic}, \textit{PickScore}, and \textit{QwenVL-7B} scores. For multimodal generation, we additionally include a \textit{CLIP-image} consistency score computed against the reference image.
We utilize QwenVL-7B as a unified vision-language reward model to assess generation quality across T2I and multimodal settings.

Our GRPO-based fine-tuning effectively bridges autoregressive modeling and multimodal RL. By integrating structured, interpretable reward signals, we enable direct policy optimization toward higher visual quality and semantic alignment. 

\noindent\textbf{(1) Text-to-Image Evaluation Prompt}

\begin{tcolorbox}[colback=gray!2!white, colframe=black!15, boxrule=0.3pt, arc=1pt, left=1pt, right=1pt, top=1pt, bottom=1pt]
\footnotesize
\textbf{Instruction:} You are a strict \textbf{reward model} for text-to-image (T2I) evaluation.  
Your goal is to produce an objective visual description and four quantitative sub-scores, with a total score in [0,4].

\textbf{Inputs:}
\begin{itemize}[leftmargin=1em]
  \item Text prompt: \texttt{"\{PROMPT\}"}
  \item Generated image: \texttt{<image>}
\end{itemize}

\textbf{Evaluation Procedure:}
\begin{enumerate}[leftmargin=1em]
  \item \textbf{Describe the generated image only} — objects, colors, lighting, composition, perspective, and visible defects. Do \emph{not} infer from the prompt.
  \item Provide four sub-scores (each $\in$ [0,1.0]):  
  (A) Prompt/Category Alignment,  
  (B) Completeness \& Composition,  
  (C) Realism \& Visual Quality,  
  (D) Artifact/Defect Freedom.
  \item Compute the final score:  
  \( \text{raw} = A + B + C + D \);  
  if any sub-score $\leq$ 0.3, cap final = 1.0. Otherwise, final = raw (rounded to two decimals).
\end{enumerate}

\textbf{Output:} JSON with fields  
\texttt{\{"description": "...", "scores": \{"A":..,"B":..,"C":..,"D":..\}, "final": ...\}}
\end{tcolorbox}

\noindent\textbf{(2) Multimodal Generation Evaluation Prompt}

\begin{tcolorbox}[colback=gray!2!white, colframe=black!15, boxrule=0.3pt, arc=1pt, left=1pt, right=1pt, top=1pt, bottom=1pt]
\footnotesize
\textbf{Instruction:} You are a strict \textbf{reward model} for subject-driven image generation with a reference image.  
Use the full [0,5] range and emphasize identity preservation and artifact freedom.

\textbf{Inputs:}
\begin{itemize}[leftmargin=1em]
  \item Reference image: \texttt{<image>}
  \item Subject name: \texttt{"\{SUBJECT\}"}
  \item Text prompt: \texttt{"\{PROMPT\}"}
  \item Generated image: \texttt{<image>}
\end{itemize}

\textbf{Evaluation Procedure:}
\begin{enumerate}[leftmargin=1em]
  \item Describe the generated image only, without inference.  
  \item Provide three sub-scores:  
  (A) Instruction Following (0–0.6),  
  (B) Identity Preservation vs.~Reference (0–3.2),  
  (C) Artifact/Defect Freedom (0–1.2).  
  \item Compute the final score:  
  \( \text{raw} = A + B + C \);  
  if A $\leq$ 0.1, B $\leq$ 0.3, or C $\leq$ 0.3, cap final = 1.0. Otherwise, final = raw (rounded to two decimals).
\end{enumerate}

\textbf{Output:} JSON with fields  
\texttt{\{"description": "...", "scores": \{"A":..,"B":..,"C":..\}, "final": ...\}}
\end{tcolorbox}

\begin{figure*}[t]
\centering
\includegraphics[width=1.0\textwidth]{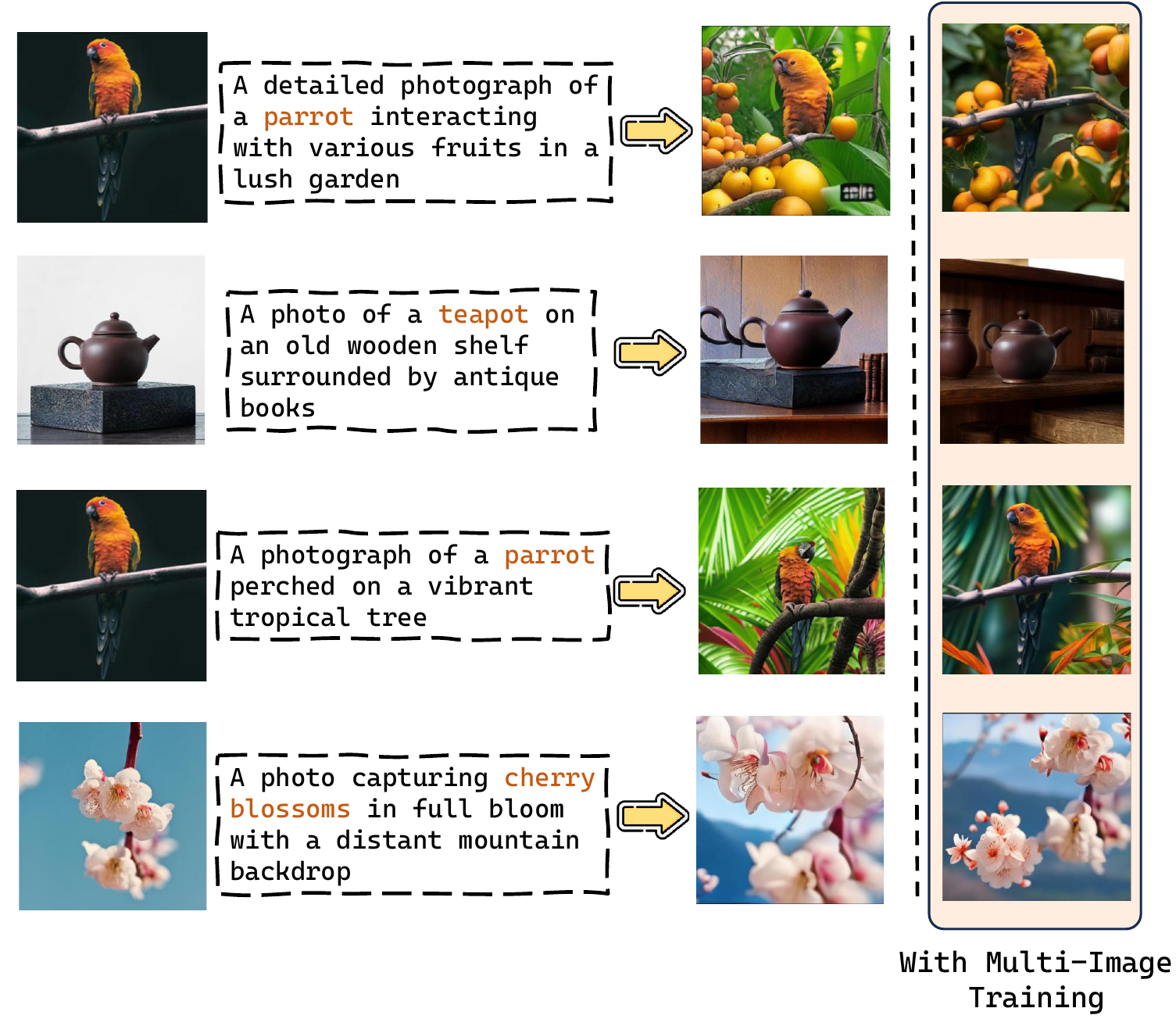}
\caption{Qualitative assessment demonstrating improved preservation of visual details by \model following multi-image training.}
\label{fig:multi}
\end{figure*}

\section{Text-to-Image Generation Evaluation}
\label{sup:t2i}
\begin{table*}[t]
    \centering
    \caption{GenEval benchmark results for text-to-image generation, classifying methods as either autoregressive or diffusion-based models. Due to our method's model size and suboptimal generators, we experience poor performance in text-to-image generation.}
    \resizebox{\textwidth}{!}{
    \begin{tabular}{@{}cl
        >{\centering\arraybackslash}m{1.4cm}
        >{\centering\arraybackslash}m{1.4cm}
        >{\centering\arraybackslash}m{1.4cm}
        >{\centering\arraybackslash}m{1.4cm}
        >{\centering\arraybackslash}m{1.4cm}
        >{\centering\arraybackslash}m{1.4cm}
        >{\centering\arraybackslash}m{1.4cm}
        @{}}
        \toprule
         & \textbf{Method} & \textbf{Single Object} & \textbf{Two Object} & \textbf{Counting} & \textbf{Colors} & \textbf{Position} & \textbf{Attribute Binding} & \textbf{Overall} \\
        \midrule

        \multirow{7}{*}{\rotatebox{90}{\textit{Autoregressive}}}
        & Chameleon~\cite{2024Chameleon}  & - & - & - & - & - & - & $0.39$ \\
        & LWM~\cite{2024LWM}  & $0.93$ & $0.41$ & $0.46$ & $0.79$ & $0.09$ & $0.15$ & $0.47$ \\
        & LlamaGen~\cite{2024llamagen} & $0.71$ & $0.34$ & $0.21$ & $0.58$ & $0.07$ & $0.04$ & $0.32$ \\
        & Show-o~\cite{2024Showo}  & $0.95$ & $0.52$ & $0.49$ & $0.82$ & $0.11$ & $0.28$ & $0.53$ \\
        & Emu$3$-Gen~\cite{2024emu3} & $0.98$ & $0.71$ & $0.34$ & $0.81$ & $0.17$ & $0.21$ & $0.54$ \\
        & Janus~\cite{2024Janus} & $0.97$ & $0.68$ & $0.30$ & $0.84$ & $0.46$ & $0.42$ & $0.61$ \\ 
        & \model & $0.87$ & $0.47$ & $0.16$ & $0.65$ & $0.11$ & $0.17$ & $0.40$ \\
        
        \midrule
        
        \multirow{12}{*}{\rotatebox{90}{\textit{Diffusion}}} 
        & LDM~\cite{2022LDM}  & $0.92$ & $0.29$ & $0.23$ & $0.70$ & $0.02$ & $0.05$ & $0.37$ \\
        & SDv$1.5$~\cite{2022LDM}  & $0.97$ & $0.38$ & $0.35$ & $0.76$ & $0.04$ & $0.06$ & $0.43$ \\
        & PixArt-$\alpha$~\cite{2023Pixelartalpha} & $0.98$ & $0.50$ & $0.44$ & $0.80$ & $0.08$ & $0.07$ & $0.48$ \\
        & SDv$2.1$~\cite{2022LDM} & $0.98$ & $0.51$ & $0.44$ & $0.85$ & $0.07$ & $0.17$ & $0.50$ \\
        & DALL-E~2~\cite{2022DALLE2} & $0.94$ & $0.66$ & $0.49$ & $0.77$ & $0.10$ & $0.19$ & $0.52$ \\
        & SDXL~\cite{2023SDXL} & $0.98$ & $0.74$ & $0.39$ & $0.85$ & $0.15$ & $0.23$ & $0.55$ \\
        & DALL-E~3~\cite{2023dalle3} & $0.96$ & $0.87$ & $0.47$ & $0.83$ & $0.43$ & $0.45$ & $0.67$ \\
        & SDv3 Medium~\cite{2024SD3} & $0.98$ & $0.74$ & $0.63$ & $0.67$ & $0.34$ & $0.36$ & $0.62$ \\
        & Flux.1 Dev~\citep{flux} & $0.98$ & $0.81$ & $0.74$ & $0.79$ & $0.22$ & $0.45$ & $0.66$ \\
        & Dream Engine~\citep{dreamengine} & $1.00$ & $0.94$ & $0.64$ & $0.81$ & $0.27$ & $0.49$ & $0.69$ \\

        \bottomrule
    \end{tabular}}
    \label{tab:exp-geneval}
\end{table*}

We evaluate the performance of our model on text-to-image (T2I) generation using the GenEval~\citep{2024Geneval} benchmarks. Results are reported in \Cref{tab:exp-geneval}.

Since \model is built upon LLaMaGen—a relatively weaker autoregressive generator—its standalone T2I performance is inferior to earlier diffusion-based models such as LDM and SDv1.5. This is expected, as models based on more advanced generators (e.g., SDXL, SD3) such as KOSMOS-G and Dream Engine consistently outperform ours in conventional T2I metrics.

Nevertheless, \model demonstrates strong performance in multimodal image generation tasks. Thanks to our proposed autoregressive architecture and a two-stage multimodal-conditioned tuning strategy, \model effectively integrates both visual and textual modalities during generation. This synergistic design compensates for its weaker generation core, enabling \model to surpass more powerful T2I models in multimodal settings, as shown in \Cref{tab:comparison}. We anticipate that incorporating stronger base generators will further improve performance. Despite its current limitations, our results suggest that \model presents a promising and efficient alternative to diffusion-based methods in multimodal scenarios.


\section{Data Construction and Formation}
\label{sec:data_const}

\begin{table*}[!htbp]
\centering
\caption{Details on dataset used in the two-stage training.}
\label{tab:dataset1}
\small 
\begin{tabular}{@{}cllc@{}}
\toprule
\textbf{Stage} & \textbf{Data Source} & \textbf{Task} & \textbf{Number of Samples} \\ 
\midrule
\multirow{3}{*}{1} 
    & Midjourney\citep{midjourney-niji-1m-llavanext} & Text to Image Generation & 700k \\ 
    & Midjourney\citep{midjourney-niji-1m-llavanext} & Image Reconstruction & 180k \\ 
    & Synthetic Data & Object Segmentation & 1.6M \\ 
\midrule
\multirow{4}{*}{2} 
    & Midjourney\citep{midjourney-niji-1m-llavanext}, Synthetic Data & Text to Image Generation & 600k \\ 
    & Synthetic Data & Object Segmentation & 150k \\ 
    & Synthetic Data, CC12M\citep{changpinyo2021cc12m} & Image Recovery & 150k \\ 
    & Subject200k\citep{OminiControl} & Subject-driven Generation & 400k \\ 
\bottomrule
\end{tabular}
\end{table*}

\paragraph{Data Formation}
Table~\ref{tab:dataset1} summarizes the datasets utilized in our two-stage training framework. Each stage is designed to progressively enhance distinct capabilities of the model using a diverse collection of multimodal data sources. In total, approximately 3 million samples are employed, with Stage 1 comprising around 2.5 million samples and Stage 2 involving 1.3 million samples, including an overlap of roughly 800k examples.

The dataset is constructed from a combination of open-source resources, such as Midjourney~\citep{midjourney-niji-1m-llavanext} and CC12M~\citep{changpinyo2021cc12m}, along with synthetic data generated via publicly available text-to-image (T2I) models, including Flux.1~\citep{flux} and Stable Diffusion v3.5~\citep{2024SD3}.

\textbf{Stage 1} focuses on establishing foundational multimodal alignment capabilities. Specifically, it includes 700k T2I samples from Midjourney~\citep{midjourney-niji-1m-llavanext}, 180k image reconstruction samples also from Midjourney, and 1.6M object segmentation samples generated through our pipeline.

\textbf{Stage 2} fine-tunes the model with 1.3 million samples. This includes 600k T2I samples—200k from Midjourney and 400k synthesized using open-source T2I models such as Flux.1~\citep{flux} and Stable Diffusion v3.5~\citep{2024SD3}. Additionally, we include 150k object segmentation samples and 150k image recovery samples, all derived from synthetic data using segmentation masks. Background images for the image recovery task are randomly selected from CC12M~\citep{changpinyo2021cc12m}.

We further incorporate 400k subject-driven image generation samples from Subject200k~\citep{OminiControl}. These samples are re-captioned using Qwen2-VL~\citep{Qwen2vl} to extract subject-relevant text and generate comprehensive image descriptions. To enrich the training set, we reverse the input-output image pairs, effectively doubling the usable data to 400k samples.

\begin{figure*}[t]
\centering
\includegraphics[width=1.0\textwidth]{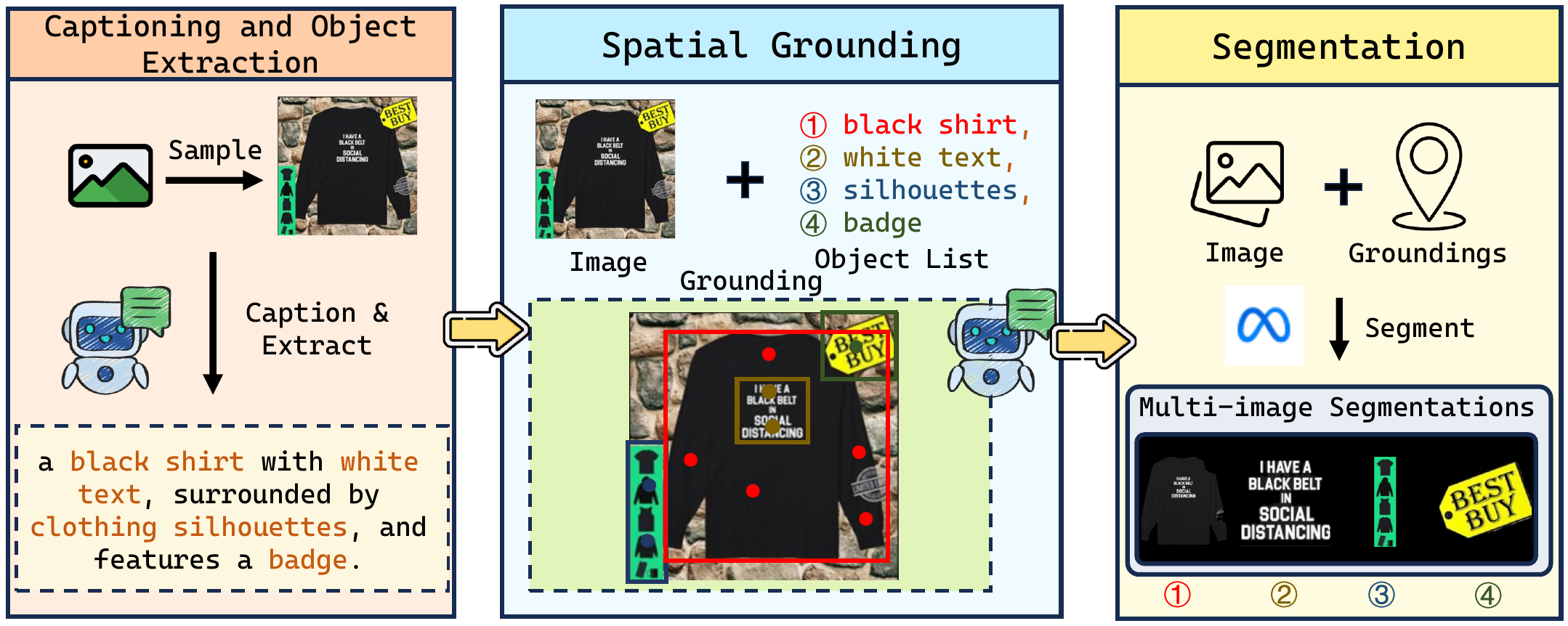}
\caption{Illustration of the automatic data generation pipeline.}
\label{fig:construction}
\end{figure*}

\paragraph{Data Construction}
To support the large-scale training required for our two-stage paradigm, we developed an automated pipeline for generating high-quality multimodal training data, as shown in Figure~\ref{fig:construction}. This pipeline combines open-source image datasets with state-of-the-art vision-language models (VLMs) and segmentation models, enabling the construction of richly annotated image-text pairs with multiple segmented foreground objects without manual labeling:

\begin{itemize}[left=2pt]
\item \textbf{Captioning and Object Extraction:} A VLM is queried to generate a comprehensive caption describing prominent elements in the image, followed by extracting a list of concrete, distinct, and segmentable objects. This ensures that the generated data focus on tangible visual entities.
\item \textbf{Spatial Grounding:} For each extracted object, the VLM is queried again to identify its spatial location within the image, returning both a tight bounding box and several representative 2D keypoints. These spatial cues constrain the region of interest for subsequent segmentation, improving accuracy and reducing background noise.
\item \textbf{Segmentation:} A segmentation model is employed to extract object masks from the image, guided by the generated bounding boxes and keypoints. This step produces high-quality masks that are both semantically aligned with the object labels and spatially accurate.
\end{itemize}

By applying this automated process to a large corpus of open-source images, we construct a diverse multimodal dataset comprising captioned images annotated with multiple precisely segmented objects. This dataset forms a critical component of our training setup, particularly enabling the object segmentation and image recovery tasks in our training paradigm.

\section{Experiment Details}
\label{sec:Experiment_Details}

\subsection{Benchmark Details}
\label{sec:Benchmark_Details}

\paragraph{DreamBench++}
\label{app:DreamBench_Plus}

\textbf{Data Organization.}
DreamBench++~\citep{peng2025dreambenchpp} comprises 150 high-quality reference images, sourced from Unsplash, Rawpixel, and Google Images, encompassing a balanced mix of subjects. These are evenly divided into three broad categories: \textit{objects}, \textit{living subjects} (humans and animals), and \textit{styles} (illustrative, painterly, etc.), ensuring visual and conceptual diversity.

In total, DreamBench++ offers \textbf{1,350 prompts} ($150 \times 9$), representing a substantial scale-up over the original DreamBench (30 subjects $\times$ 25 prompts). Relative to DreamBench, the dataset is \textbf{5$\times$ larger in subjects} and \textbf{54$\times$ larger in prompts}, enabling broader evaluation of generative performance.

\textbf{Evaluation Metric.}
DreamBench++ adopts an automatic, GPT-4o-based evaluation protocol designed to closely mirror human judgment. Each generated image is assessed against both its reference image and its corresponding prompt, using two complementary axes:

\begin{itemize}[left=2pt, itemsep=0.5pt,topsep=0.5pt]
\item \textbf{Concept Preservation (CP):} Measures fidelity between the generated image and the reference. Key attributes include shape, color, texture, and facial details.
\item \textbf{Prompt Following (PF):} Evaluates how well the generation aligns with the prompt in terms of relevance, accuracy, completeness, and contextual appropriateness.
\end{itemize}

Each axis is scored on a \textbf{five-level ordinal scale} from 0 (Very Poor) to 4 (Excellent), avoiding the complexity and bias of pairwise comparisons.

\textbf{Metric Justification.}
Following DreamBench++, we adopt \textbf{CP$\cdot$PF} as the primary score for balanced performance and \textbf{CP/PF} as an indicator of modality overfitting.
\textit{Why CP$\cdot$PF instead of CP+PF?}
The additive form can assign high scores to models excelling on only one modality (e.g., CP$\approx$1, PF$\approx$0.1 still yields a large sum). In contrast, CP$\cdot$PF approaches zero when \textit{either} metric is poor, thus more sensitively reflecting imbalanced multimodal control.
\textit{Why CP/PF as an overfitting indicator?}
On DreamBench++, the dominant failure mode is \textit{image-dominant} behavior---copying the reference while ignoring text. CP/PF$>$1 directly captures this (e.g., Lumina-mGPT and Unified-IO2 have CP/PF$>$3). Using PF/CP instead would flag text-dominant failures, which are less common in this benchmark.

\begin{figure*}[t]
\centering
\includegraphics[width=0.85\textwidth]{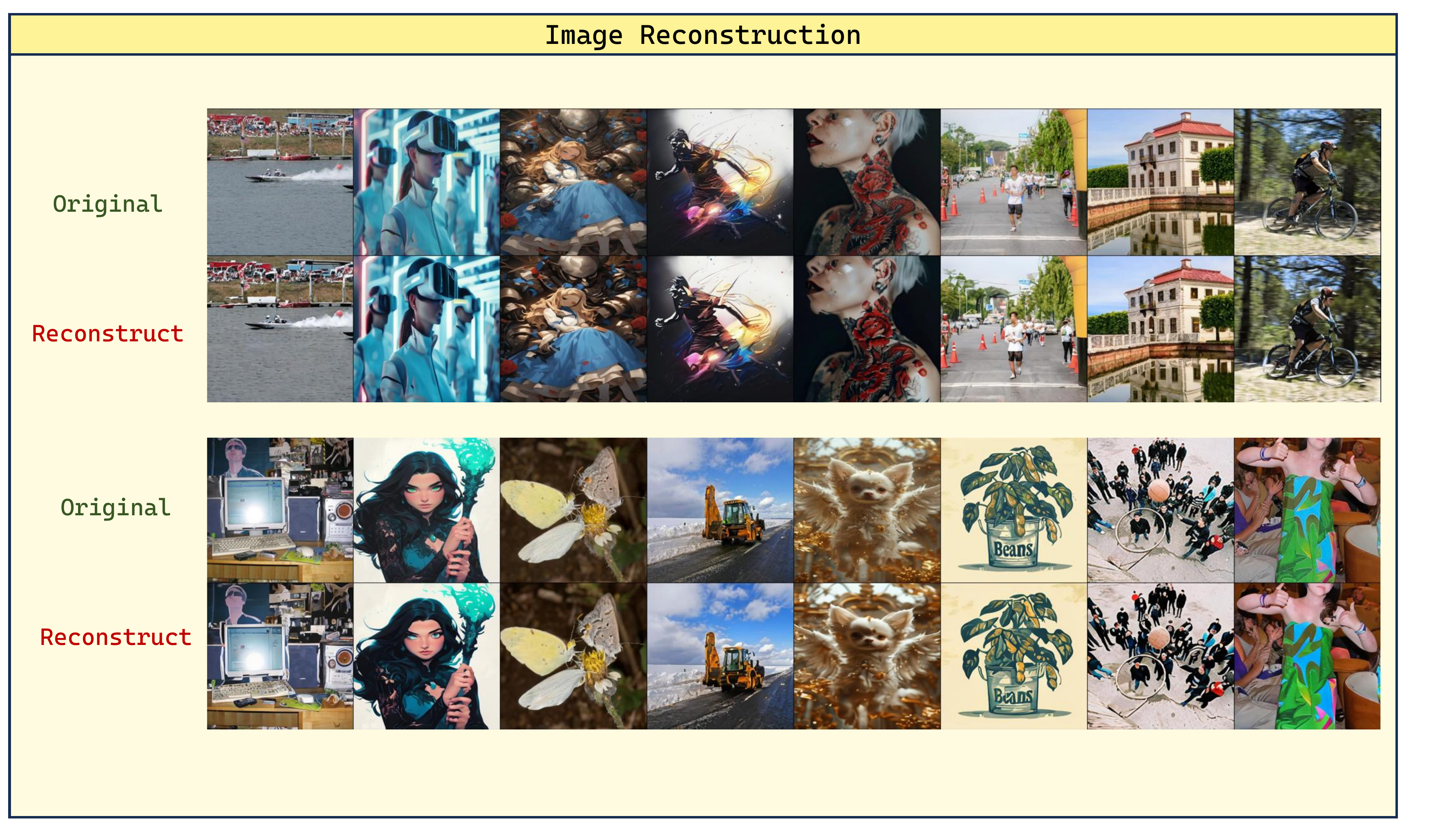}
\caption{Qualitative comparison of image reconstruction results using \model.}
\label{fig:reconstruction}
\end{figure*}

\begin{figure*}[t]
\centering
\includegraphics[width=0.85\textwidth]{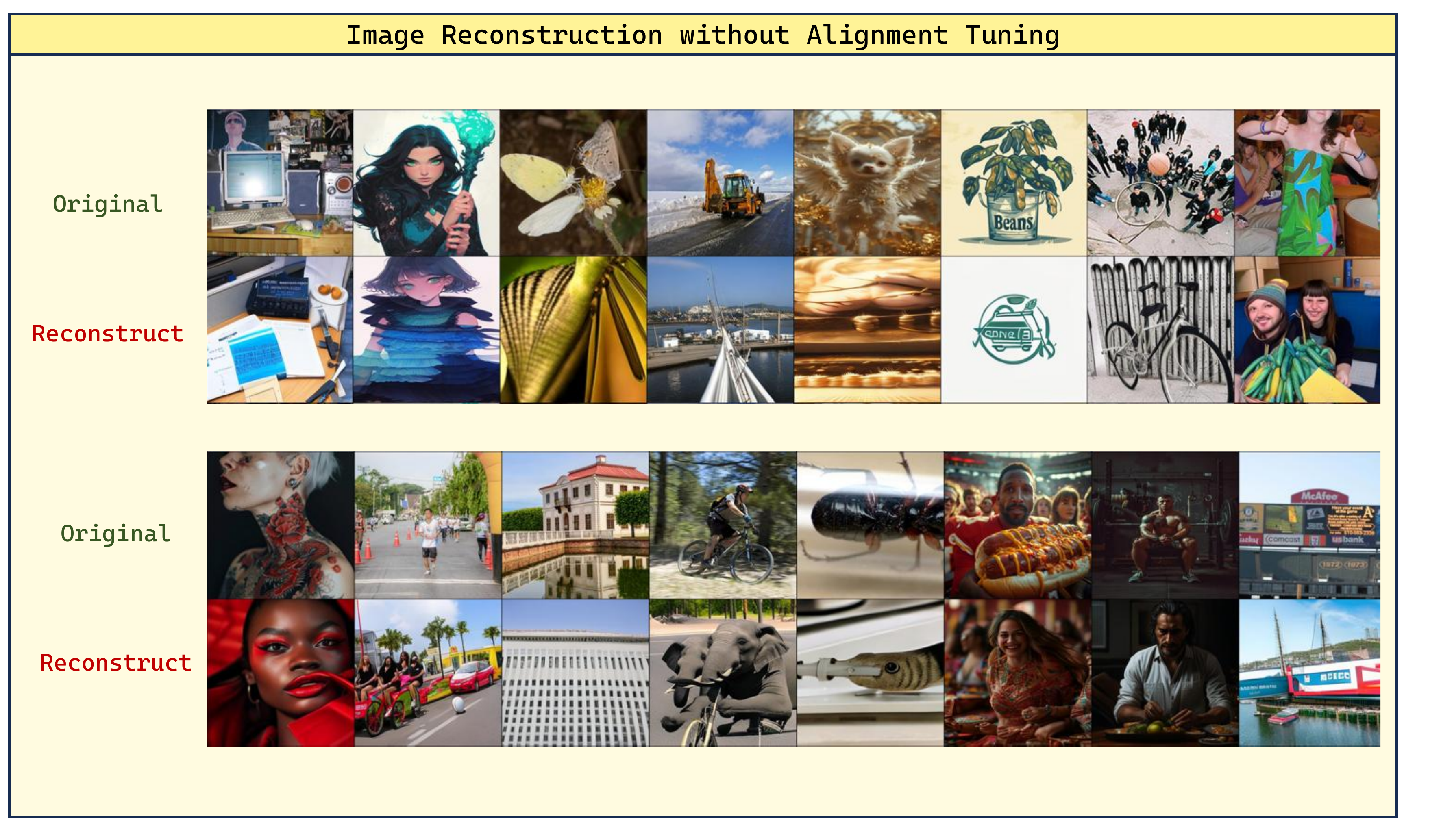}
\caption{Image reconstruction results of \model \textit{without} alignment tuning.}
\label{fig:reconstruction_wo_align}
\end{figure*}
\paragraph{DreamBench}
The original DreamBench~\citep{ruiz2023dreamboothfinetuningtexttoimage} dataset consists of 30 subjects, each paired with 25 prompts, totaling 750 prompt-image pairs. It serves as a foundational benchmark for evaluating personalized image generation models, focusing on the model's ability to maintain subject identity across diverse prompts.

\subsection{Baselines}
\label{sec:Baselines_Details}

We compare our method against various baselines, categorized as follows:

\begin{itemize}[left=2pt, itemsep=0.5pt,topsep=0.5pt]
\item \textbf{Textual Inversion}~\citep{gal2022imageworthwordpersonalizing} learns a new word embedding to represent a specific concept, enabling personalized image generation by incorporating the new token into prompts. It requires a few images of the subject and fine-tunes the embedding without altering the base model weights.
\item \textbf{DreamBooth}~\citep{ruiz2023dreamboothfinetuningtexttoimage}: DreamBooth fine-tunes a pre-trained text-to-image model to bind a unique identifier with the subject's visual concept, allowing for personalized generation. It requires several images of subject and modifies model weights to capture subject-specific details.

\item \textbf{BLIP-Diffusion}~\citep{li2023blipdiffusionpretrainedsubjectrepresentation}: This approach introduces a pre-trained multimodal encoder to provide subject representations for the diffusion generator, enabling controllable multimodal image generation. 

\item \textbf{KOSMOS-G}~\citep{Kosmos-G}: KOSMOS-G is a multimodal large language model designed for zero-shot image generation from interleaved vision-language inputs, including multiple images and text. It aligns the output space of a transformer-based causal language model with a diffusion-based image decoder using a lightweight AlignerNet and compositional instruction tuning. This architecture enables KOSMOS-G to perceive complex multimodal prompts and generate coherent, subject-driven images without modifying the base image decoder.

\item \textbf{Emu2}~\citep{emu2}: Emu2 is a 37-billion-parameter generative multimodal model trained on large-scale multimodal sequences with a unified autoregressive objective. It exhibits strong in-context learning abilities for various multimodal tasks, including visual prompting and object-grounded generation.

\item \textbf{IP-Adapter}~\citep{ye2023ip-adapter}: IP-Adapter is a lightweight adapter that enables image prompt capability for pre-trained text-to-image diffusion models. It integrates image features into the generation process without modifying the base model, supporting flexible and efficient image-to-image generation.

\item \textbf{DreamEngine}~\citep{dreamengine}: DreamEngine is a unified framework that integrates multimodal encoders with diffusion models through a two-stage training approach, enabling advanced text-image interleaved control and achieving state-of-the-art performance in generating images with complex, concept-merged inputs.

\item \textbf{Unified-IO 2}~\citep{lu2023unifiedio2}: Unified-IO 2 is an autoregressive multimodal model capable of understanding and generating images, text, audio, and actions. It tokenizes various modalities into a shared semantic space and processes them with a single encoder-decoder transformer. Trained from scratch on a large multimodal pre-training corpus and fine-tuned on an ensemble of 120 datasets, Unified-IO 2 achieves state-of-the-art performance on the GRIT benchmark and strong results across more than 35 benchmarks. 

\item \textbf{Lumina-mGPT}~\citep{2024lumina}: Lumina-mGPT is a multimodal autoregressive model designed for flexible photorealistic text-to-image generation. It employs a pretrained decoder-only transformer as a unified framework for modeling multimodal token sequences. Through multimodal Generative PreTraining (mGPT) and subsequent Flexible Progressive Supervised Finetuning (FP-SFT) and Omnipotent Supervised Finetuning (Omni-SFT), Lumina-mGPT demonstrates versatile multimodal capabilities, including visual generation tasks, controllable generation tasks and vision-language tasks. 

\end{itemize}

\section{Qualitative Study}
\label{sec:Qualitative_Study}

\subsection{Image Reconstruction}
\label{sec:Reconstruction}

As illustrated in Figures~\ref{fig:reconstruction} and \ref{fig:reconstruction_wo_align}, \model demonstrates strong image reconstruction capabilities following two-stage training. Notably, it is able to effectively reconstruct input images and preserve fine-grained visual details, even when input images are of low resolution (224×224) and outputs are generated at 512×512 resolution.

In contrast, when alignment tuning is omitted, although the model benefits from the pretrained multimodal encoder and the proposed architecture, it tends to treat the input image as a visual prompt akin to a caption. As shown in Figure~\ref{fig:reconstruction_wo_align}, this leads to outputs that resemble descriptive interpretations of the input rather than faithful reconstructions. Consequently, visual fidelity and spatial consistency degrade significantly without alignment tuning.

\subsection{Text-guided Image Segmentation}
\label{sec:Segmentation}

\begin{figure*}[t]
\centering
\includegraphics[width=1.0\textwidth]{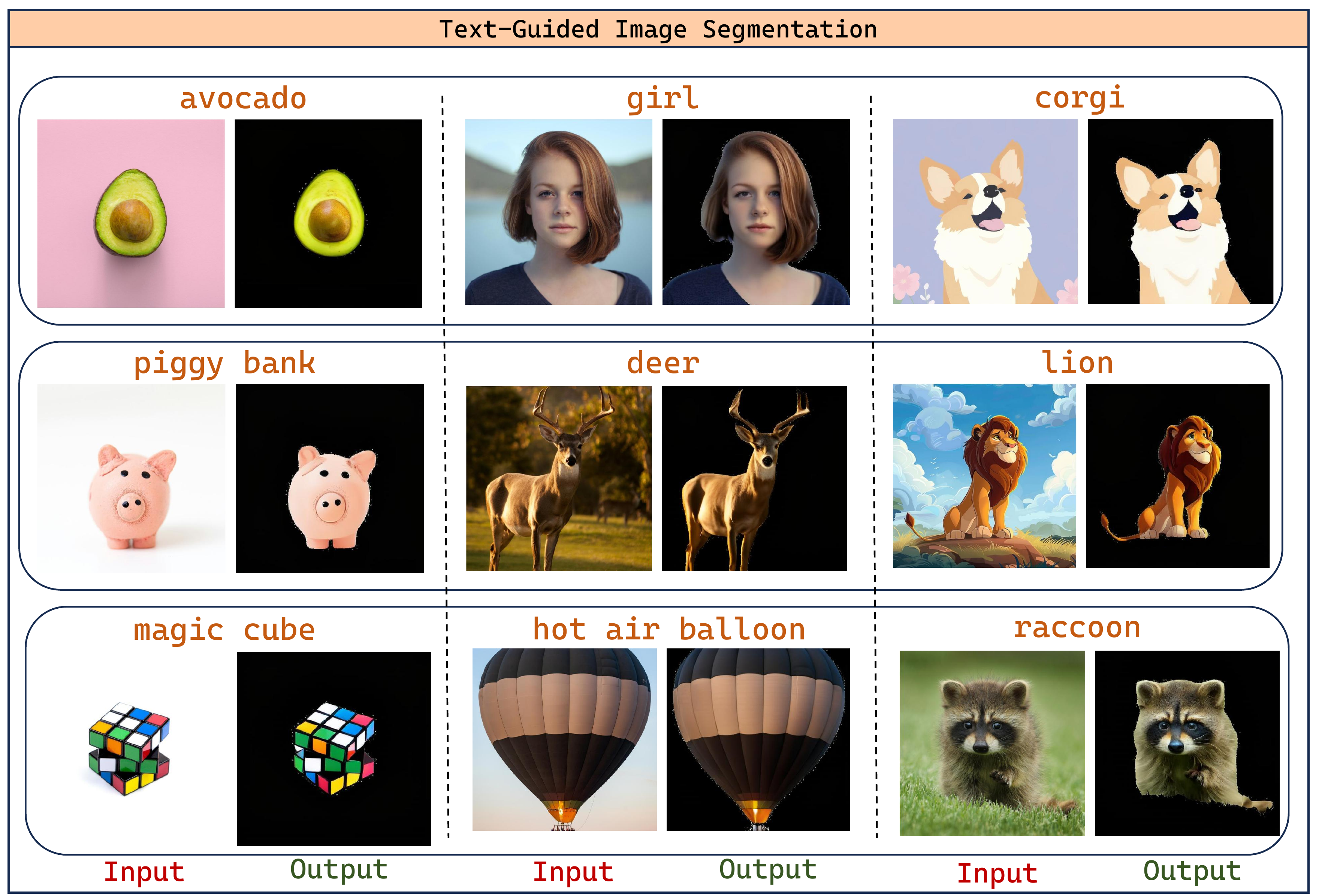}
\caption{Qualitative results of text-guided image segmentation using \model.}
\label{fig:segmentation_example}
\end{figure*}

We evaluate \model on the DreamBench\texttt{++} benchmark to assess its performance in text-guided image segmentation. As demonstrated in Figure~\ref{fig:segmentation_example}, \model successfully identifies and segments visual concepts corresponding to the given textual prompts. These results highlight the model's ability to generalize across tasks and showcase its robust multimodal understanding and generation.

\paragraph{Open-Vocabulary Segmentation.}
To further assess segmentation capability, we evaluate \model on ADE20K (A-150), a standard open-vocabulary segmentation benchmark. Notably, we perform zero-shot evaluation directly after Stage-1 pretraining, without any task-specific fine-tuning on these datasets. Due to the fixed output resolution of our generator, all input images are resized to 512$\times$512 for inference.

\begin{figure*}[t]
\centering
\includegraphics[width=1.0\textwidth]{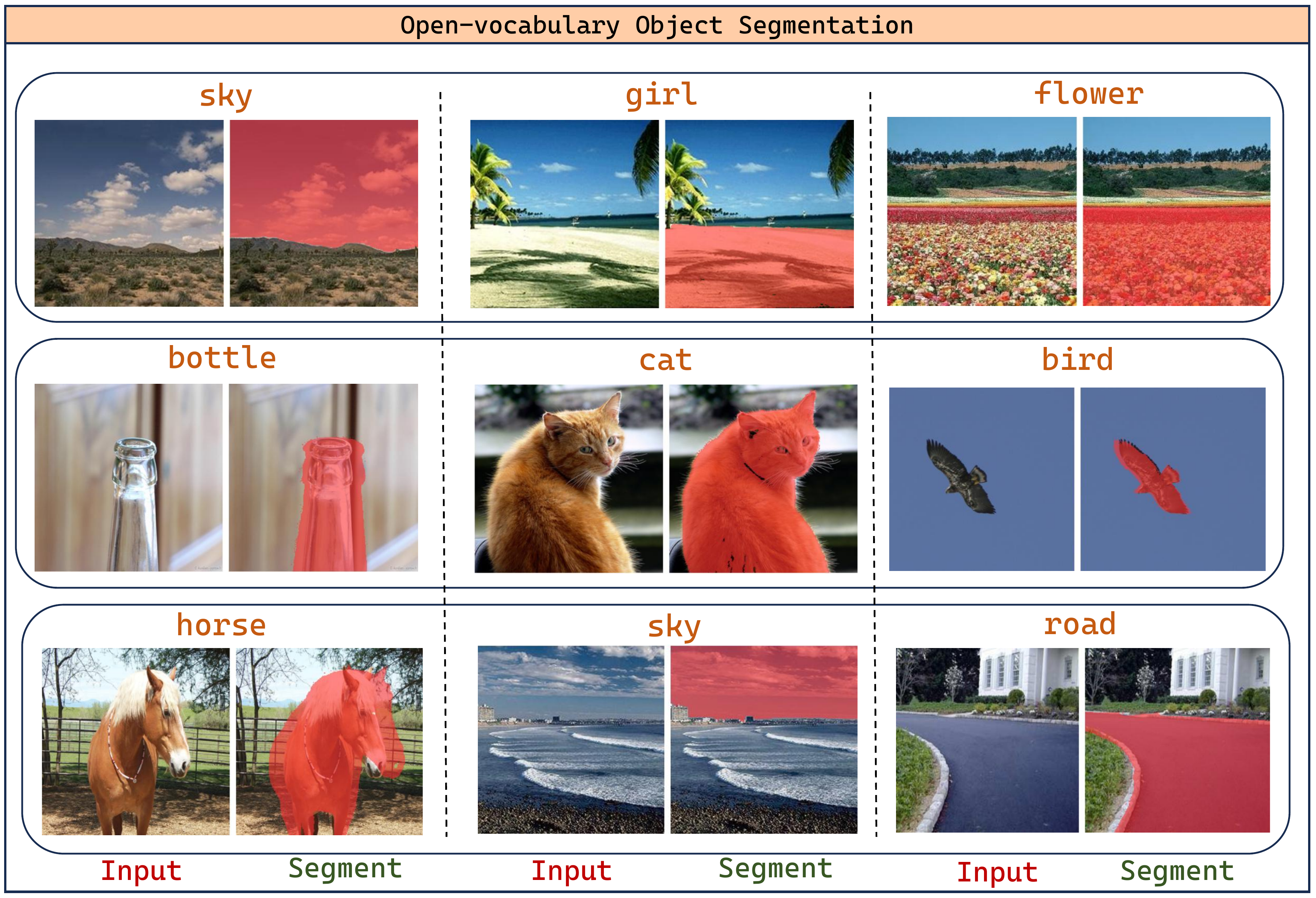}
\caption{Qualitative results of open-vocabulary segmentation using \model on ADE20K~\citep{ADE20K}, PASCAL Context~\citep{pascal_CONTEXT}, and PASCAL VOC~\citep{pascal_VOC} benchmarks. The generated segmentation masks are visualized in red for clarity.}
\label{fig:open_segmentation_example}
\end{figure*}


\begin{figure*}[t]
\centering
\includegraphics[width=0.75\textwidth]{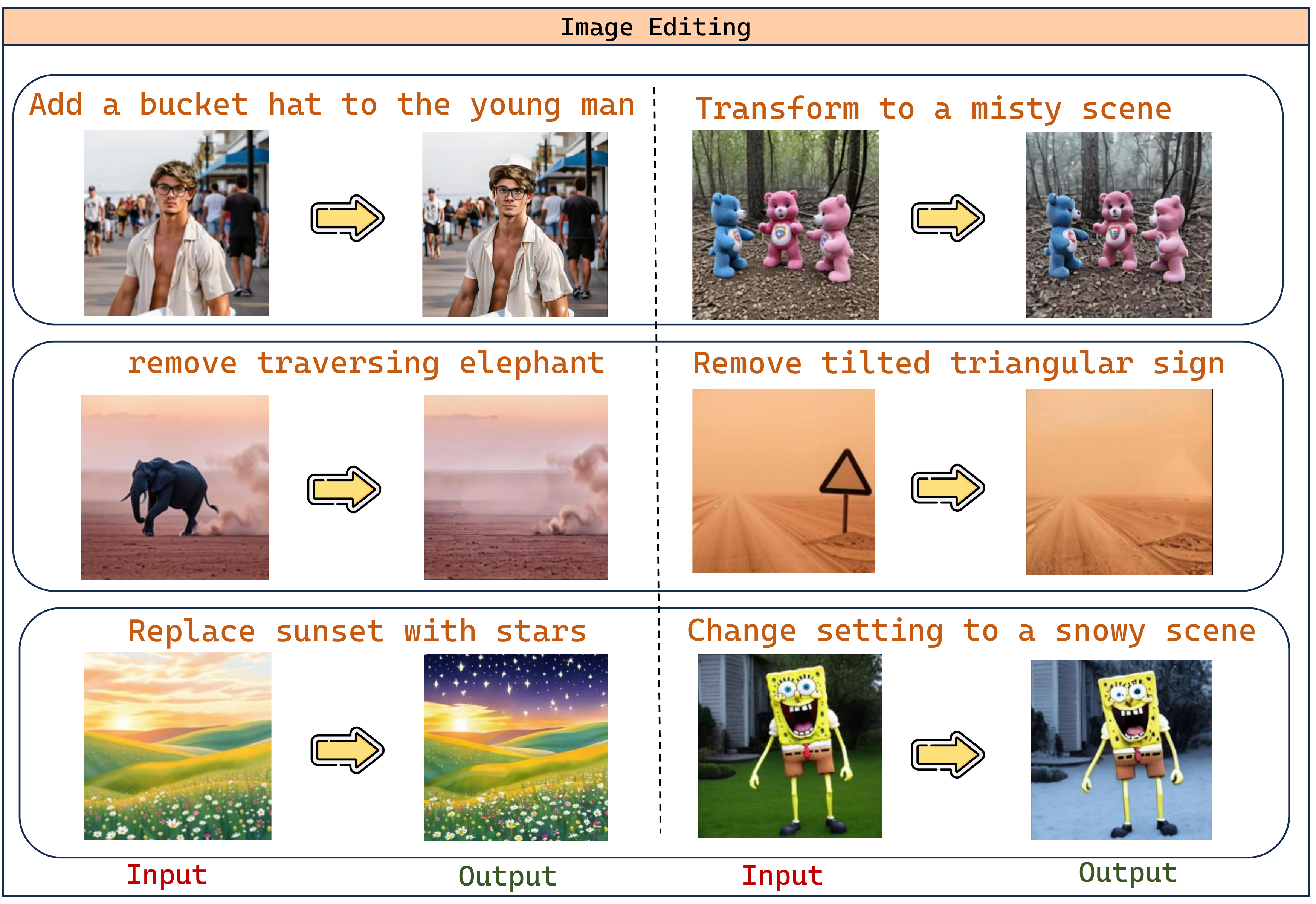}
\caption{Qualitative examples from Image Editing. The model is finetuned on 1M samples from ImgEdit~\citep{ImgEdit} and OmniEdit~\citep{OmniEdit} dataset. Although restricted by the performance of image generator, \model can generate visually consistent edits.}
\label{fig:image_edit_example}
\end{figure*}

\begin{table*}[h]
\centering
\small
\caption{Zero-shot open-vocabulary segmentation on ADE20K-150~\citep{ADE20K}. We report the mIoU score as the metric.}
\label{tab:seg_benchmark_a150}
\resizebox{1.0\textwidth}{!}{%
\begin{tabular}{lccccccc}
\toprule
\multirow{2}{*}{\textbf{Dataset}} &
LSeg+ & OpenSeg & SimBaseline & ZegFormer & SimSeg & DreamLIP & \multirow{2}{*}{\model} \\
& \citep{LSeg} & \citep{LSeg} & \citep{LSeg} & \citep{LSeg} & \citep{LSeg} & \citep{LSeg} & \\
\midrule
ADE20K-150 & 13.00 & 15.30 & 15.30 & 16.40 & 20.50 & 17.10 & 19.91 \\
\bottomrule
\end{tabular}
}
\end{table*}

As shown in Table~\ref{tab:seg_benchmark_a150} and Figure~\ref{fig:open_segmentation_example}, \model achieves competitive performance on the benchmark compared with the traditional open-vocabulary segmentation methods in a zero-shot manner. Compared with these methods that require massive task-specific pretraining, \model shows competitive performance after Stage-1 training in a zero-shot manner.
These results demonstrate that our two-stage training effectively learns to balance visual and textual guidance, enabling the model to generate accurate segmentation masks for novel categories. While performance is constrained by the current backbone capacity and data scale, these findings highlight the extensibility and promising potential of our framework for broader vision-language tasks.

\subsection{Multi-Image Generation}
\label{sec:Multi-Image}

\begin{figure*}[t]
\centering
\includegraphics[width=0.85\textwidth]{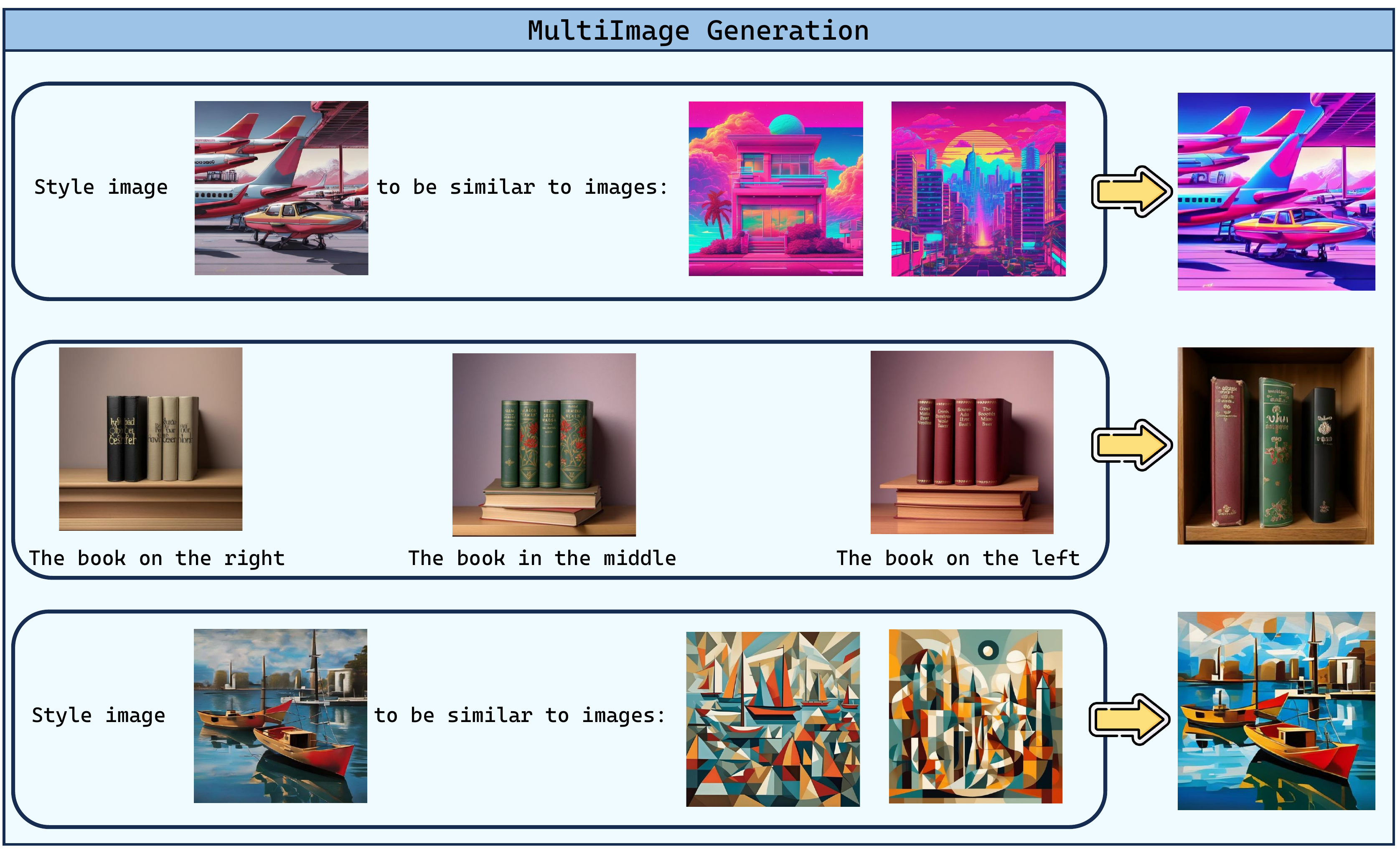}
\caption{Qualitative results for multi-image generation.}
\label{fig:multi_image}
\end{figure*}

We evaluate \model on multi-image generation tasks using the X2I dataset~\citep{OmniGen}. As shown in Figure~\ref{fig:multi_image}, the model is capable of generating visually consistent outputs conditioned on the multi-image inputs. The generated images reflect coherent semantics, style, and layout across the samples.

\subsection{Multimodal In-Context Image Generation}
\label{sec:mmicl}

\begin{figure*}[t]
\centering
\includegraphics[width=0.75\textwidth]{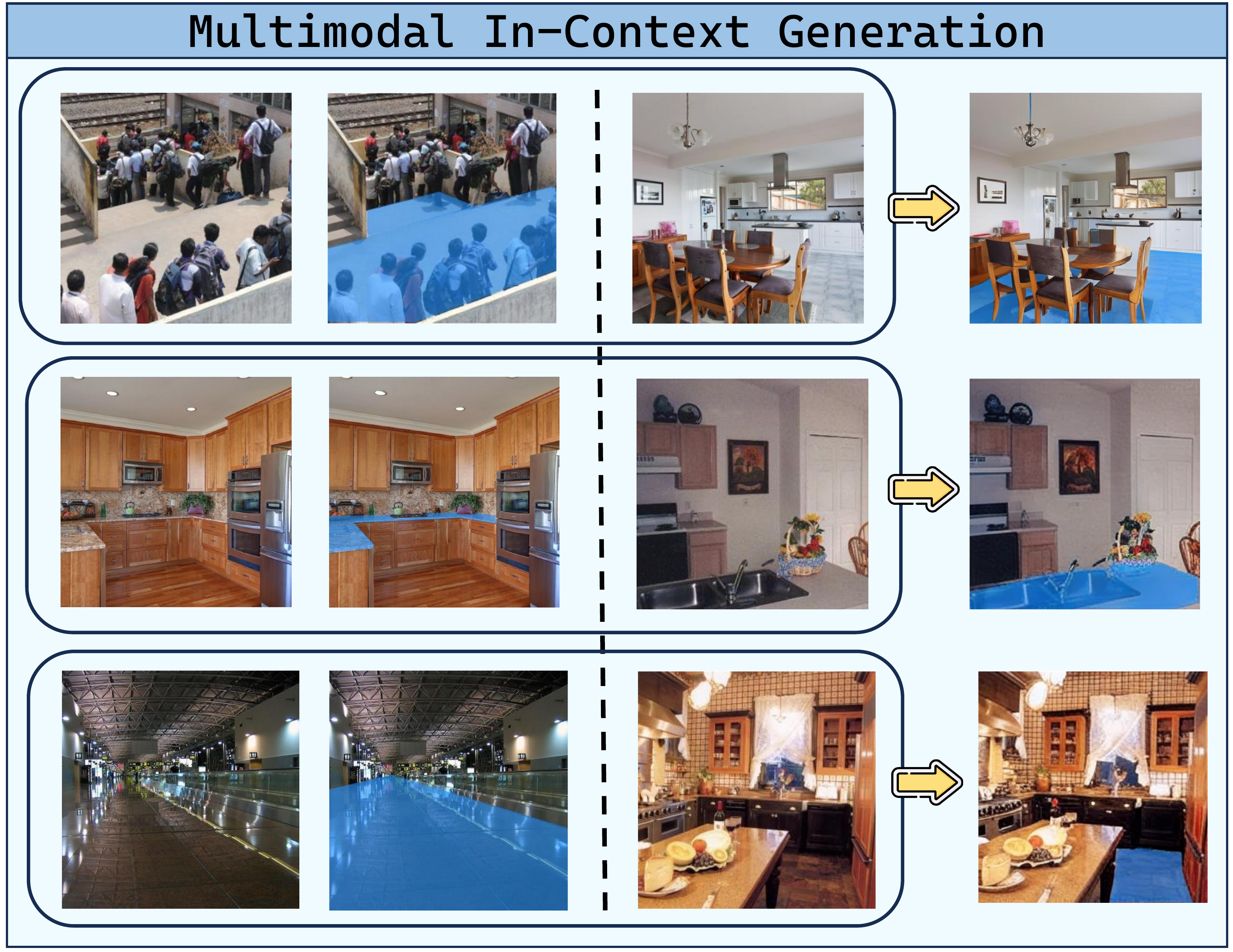}
\caption{Qualitative examples from multimodal in-context image generation. The model adapts to patterns in the visual context.}
\label{fig:icl_example}
\end{figure*}

To assess \model's few-shot generalization capabilities, we evaluate it on the multimodal in-context image generation task using the X2I-ICL dataset~\citep{OmniGen}. As illustrated in Figure~\ref{fig:icl_example}, \model learns to synthesize images that follow the stylistic patterns demonstrated in the in-context examples. This indicates its capability to infer complex visual trends and align generation with image context.

\section{Versatility Across Different Multimodal Tasks}
\label{app:applications}

To assess the broad applicability of our proposed framework, we evaluate \model across a diverse set of multimodal generation tasks, including text-guided image segmentation, subject-driven image generation, multi-image generation, and multimodal in-context learning. For each task, we apply supervised fine-tuning where necessary, ensuring robust generalization while maintaining architectural consistency.

\paragraph{Image Segmentation.}
We evaluate this task directly after Stage 1 training, without additional fine-tuning. The model demonstrates strong object localization and mask precision from prompt-aligned inputs, confirming the effectiveness of the proposed training pipeline and segmentation-aware data construction process.

\paragraph{Subject-driven Image Generation.}
This task is evaluated using the model at the end of Stage 2. No additional task-specific tuning is applied. The model successfully generates high-fidelity, identity-preserving images consistent with subject descriptors.

\paragraph{Multi-Image Generation.}
We fine-tune the Stage 2 model on a subset of X2I-subject-driven~\citep{OmniGen} dataset for two additional epochs using a reduced learning rate of $5 \times 10^{-5}$. All other optimization settings remain consistent with Stage 2. The dataset is split into disjoint training and test sets, and quantitative results are reported on the test split. The model learns to generate visually diverse yet semantically aligned images for the same input.

\paragraph{Multimodal In-Context Learning.}
We fine-tune the model for 10 epochs on the X2I-ICL dataset~\citep{OmniGen}, which features sequences of input-output pairs for in-context generalization. We use a learning rate of $5 \times 10^{-5}$ and ensure a strict train-test separation. The model adapts to context examples and generates new samples following the observed patterns, showing strong in-context learning performance without explicit prompt engineering.

\paragraph{Conclusion.}
The qualitative results presented in Section~\ref{sec:Qualitative_Study} confirm the versatility of \model across a wide range of tasks. Notably, the model adapts to each task without architectural modifications, requiring only lightweight fine-tuning.

\section{Human Evaluation}
\label{app:human_eval}

To validate the reliability of GPT-based evaluation metrics used in DreamBench++, we conduct a human study on 50 randomly sampled images from our model's outputs. Three independent annotators rate each image on Concept Preservation (CP) and Prompt Following (PF) using the same five-level ordinal scale (0--4) as the automatic evaluation.

We report two complementary measures of reliability:
\begin{itemize}[leftmargin=*, itemsep=2pt, topsep=2pt]
\item \textbf{Inter-annotator agreement:} Measured by Krippendorff's $\alpha$, we obtain $\alpha = 0.76$ for CP (substantial agreement) and $\alpha = 0.62$ for PF (moderate agreement). These values indicate consistent human judgment across annotators.
\item \textbf{Human-metric consistency:} Defined as the proportion of cases where the GPT-based score matches the majority human judgment, this metric reaches 84\% for CP and 94\% for PF. These figures align closely with the consistency rates reported in DreamBench++ (83.31\% for CP, 98.17\% for PF).
\end{itemize}
These results confirm that the GPT-based metrics serve as reliable proxies for human evaluation, supporting their use as the primary evaluation protocol in our experiments.

\section{CP--PF Trade-off Analysis}
\label{app:cp_pf_tradeoff}

A central challenge in multimodal-conditioned image generation is balancing fidelity to the reference image (Concept Preservation, CP) against adherence to the text prompt (Prompt Following, PF). In this section, we conduct systematic experiments to analyze the CP--PF trade-off by varying two key factors: (1) Stage-2 task mixing ratios during training, and (2) classifier-free guidance (CFG) scale at inference.

\subsection{Effect of Task Weight Distribution}
\label{app:task_weight}

We investigate how the composition of training tasks in Stage-2 affects the balance between concept preservation and prompt following. Stage-2 comprises four task types with configurable sampling weights: text-to-image generation (T2I), object segmentation (Seg), image recovery (Rec), and subject-driven generation (Sub). We evaluate five representative weight configurations, as reported in Table~\ref{tab:task_weight}.

\begin{table*}[t]
\centering
\small
\caption{Effect of Stage-2 task weight distribution on DreamBench++. Left columns show sampling weights for each task type; right columns report evaluation metrics.}
\label{tab:task_weight}
\begin{tabular}{@{}l cccc c ccc@{}}
\toprule
& \multicolumn{4}{c}{\textbf{Task Sampling Weights}} & & \multicolumn{3}{c}{\textbf{Metrics}} \\
\cmidrule{2-5} \cmidrule{7-9}
\textbf{Configuration} & T2I & Seg & Rec & Sub & & CP$\uparrow$ & PF$\uparrow$ & CP$\cdot$PF$\uparrow$ \\
\midrule
Default (Ours) & 0.45 & 0.15 & 0.15 & 0.25 & & 0.557 & 0.840 & \textbf{0.468} \\
Image-heavy & 0.20 & 0.25 & 0.25 & 0.30 & & 0.609 & 0.741 & 0.451 \\
Text-heavy & 0.80 & 0.05 & 0.05 & 0.10 & & 0.383 & 0.915 & 0.350 \\
w/o Rec & 0.45 & 0.25 & 0.05 & 0.25 & & 0.588 & 0.666 & 0.392 \\
w/o Seg & 0.45 & 0.05 & 0.25 & 0.25 & & 0.390 & 0.880 & 0.343 \\
\bottomrule
\end{tabular}
\end{table*}

The results reveal a systematic trade-off between concept preservation and prompt following:
\begin{itemize}[leftmargin=*, itemsep=2pt, topsep=2pt]
\item \textbf{Image-heavy:} Increasing weights for image-centric tasks (Seg, Rec, Sub) improves CP from 0.557 to 0.609 but reduces PF from 0.840 to 0.741.
\item \textbf{Text-heavy:} Emphasizing T2I (weight = 0.80) yields higher PF (0.915) at the cost of significantly degraded CP (0.383).
\item \textbf{w/o Rec:} Reducing image recovery weight primarily degrades PF and induces copy-paste behavior.
\item \textbf{w/o Seg:} Minimizing segmentation weight mainly hurts CP, impairing fine-grained visual grounding.
\end{itemize}
These observations are consistent with our ablation study (\S\ref{sec:ablation}), confirming that the default configuration achieves the optimal balance as measured by CP$\cdot$PF.

\subsection{Effect of Classifier-Free Guidance Scale}
\label{app:cfg_sweep}

We examine how classifier-free guidance (CFG) scale at inference time affects the CP--PF trade-off. Table~\ref{tab:cfg_sweep} summarizes the results across four CFG values.

\begin{table}[t]
\centering
\small
\caption{Effect of CFG scale on DreamBench++ metrics. Higher CFG strengthens prompt adherence but may degrade concept preservation at extreme values.}
\label{tab:cfg_sweep}
\setlength{\tabcolsep}{6pt}
\begin{tabular}{@{}lcccc@{}}
\toprule
\textbf{CFG Scale} & \textbf{3.5} & \textbf{5.5} & \textbf{7.5} & \textbf{9.5} \\
\midrule
CP$\uparrow$ & 0.532 & 0.551 & 0.557 & 0.481 \\
PF$\uparrow$ & 0.815 & 0.828 & 0.840 & 0.881 \\
CP$\cdot$PF$\uparrow$ & 0.434 & 0.456 & \textbf{0.468} & 0.424 \\
\bottomrule
\end{tabular}
\end{table}

Increasing CFG leads to monotonic improvement in PF, as stronger guidance amplifies the influence of text conditioning. However, CP exhibits a non-monotonic pattern: it initially rises with CFG but drops sharply at CFG$=$9.5, likely due to over-amplification that distorts fine-grained visual features. The combined metric CP$\cdot$PF peaks at CFG$=$7.5, which we adopt as the default following the DreamBench++ protocol.

\end{document}